\documentclass[conference]{IEEEtran}
\IEEEoverridecommandlockouts
\usepackage{cite}
\usepackage{amsmath,amssymb,amsfonts}
\usepackage{algorithmic}
\usepackage{graphicx}
\usepackage{textcomp}
\usepackage{xcolor}
\def\BibTeX{{\rm B\kern-.05em{\sc i\kern-.025em b}\kern-.08em
    T\kern-.1667em\lower.7ex\hbox{E}\kern-.125emX}}
\usepackage{threeparttable}
\usepackage{footnote}
\makesavenoteenv{table}
\usepackage[symbol]{footmisc}
\usepackage{amsmath}
\usepackage{mathtools}

\usepackage{multirow}
\usepackage{booktabs} 

\usepackage[backref]{hyperref} 
\hypersetup{hidelinks}

\usepackage{tikz}

\usepackage{caption}
\usepackage{subcaption}

\usepackage{nicefrac}

\usepackage[numbers,sort&compress]{natbib}

\usepackage{fancyhdr}

\usepackage{graphicx}
\graphicspath{{./Figures_Tables/}}

\begin{document}

\title{Predicting Influenza A Viral Host Using PSSM and Word Embeddings}

\author{\IEEEauthorblockN{Yanhua Xu}
\IEEEauthorblockA{\textit{Department of Computer Science} \\
\textit{University of Liverpool}\\
Liverpool, UK \\
y.xu137@liverpool.ac.uk}
\and
\IEEEauthorblockN{Dominik Wojtczak}
\IEEEauthorblockA{\textit{Department of Computer Science} \\
\textit{University of Liverpool}\\
Liverpool, UK \\
d.wojtczak@liverpool.ac.uk}
}

\maketitle

\begin{abstract}
The rapid mutation of influenza virus threatens public health. Reassortment among viruses with different hosts can lead to a fatal pandemic. However, it is difficult to detect the original host of the virus during or after an outbreak as influenza viruses can circulate between different species. Therefore, early and rapid detection of the viral host would help reduce the further spread of the virus. We use various machine learning models with features derived from the position-specific scoring matrix (PSSM) and features learned from word embedding and word encoding to infer the origin host of viruses. The results show that the performance of the PSSM-based model reaches the MCC around 95\%, and the $F_1$ around 96\%. The MCC obtained using the model with word embedding is around 96\%, and the $F_1$ is around 97\%.
\end{abstract}


\section{Introduction}
\IEEEPARstart{I}{nfluenza} is an infectious disease that occurs globally and infects up to 20 percent of the world’s population each year, although its prevalence is usually underestimated~\cite{sec1_1}. Influenza pandemics occur at a lower frequency than seasonal influenza epidemics, but each such crisis can cause millions of deaths. Flu epidemics seriously impact vulnerable people with chronic medical conditions, and flu pandemics put people of all ages at life-threatening risk.

Influenza viruses are divided into four types based on their internal ribonucleoproteins: A, B, C and D. Influenza D viruses are not known to cause human illness. Influenza C viruses only affect humans, but they are less likely to cause large-scale pandemics or seasonal epidemics. Thus, currently, seasonal flu vaccine strains do not inoculate against influenza C and D viruses. Influenza A and B viruses are the major causes of seasonal epidemics. Influenza B viruses also only affect humans, but influenza A viruses affect both humans and animals and can cause global epidemics (i.e., pandemics).  Subtypes of influenza A viruses are differentiated by two kinds of glycoproteins under the viral envelope: Hemagglutinin (HA) and Neuraminidase (NA). Within these types, 18 HA subtypes (numbered 1–18) and 11 NA subtypes (numbered 1–11) have been discovered to date~\cite{sec1_2}.

The characteristic of antigenic sites on HA or NA protein that is recognized by the immune system to inhibit flu infectious changes rapidly to escape the recognition of the immune system. This process is also known as antigenic drift, which results in new influenza A, B or C virus strains that are partially recognized by humans’ immune systems and contribute to seasonal influenza outbreaks. The HA or NA in influenza A can experience drastic changes on antigenic sites and cause an antigenic shift. Antigenic shifts may result from a re-assortment of different viruses within single or multiple hosts and generate a novel virus~\cite{sec1_3}. Multiple pandemics have resulted from extreme antigenic shifts, after which most people lack immunity to the novel virus. The origins of four major influenza pandemics that emerged since 1900 may have been caused by a recombination of animal viruses (swine and avian) and human viruses: Spanish flu (1918–1919), Asian flu (1957–1958), Hong Kong flu (1968–1969) and the 2009 flu pandemic (2009–2010).

The Spanish flu was caused by the A/H1N1 virus. It is the deadliest pandemic in the recorded history and killed an estimated 17 – 100 million people~\cite{sec1_4},~\cite{sec1_5},~\cite{sec1_6}. The origin host of the Spanish flu remains a mystery~\cite{sec1_7}, but recent studies suggest it may have sprung from birds or pigs to humans~\cite{sec1_8},~\cite{sec1_9},~\cite{sec1_10}. After the initial pandemic in 1918, the virus adapted to keep playing a major role in flu epidemics until 1957, when the major changes in HA and NA produced the novel virus A/H2N2 and resulted in the Asian flu pandemic. Asian flu has higher morbidity and mortality compared with the subsequent Hong Kong flu in 1968 as Hong Kong flu caused by the A/H3N2 virus which only involved the major changes in the HA antigen~\cite{sec1_11}. Both Asian flu and Hong Kong flu were caused by reassortment between human viruses and avian viruses. The A/H1N1 virus also caused the 2009 flu pandemic, but that iteration involved a complex triple reassortment between human, avian and swine viruses~\cite{sec1_12},~\cite{sec1_13}.

Influenza viruses have multiple hosts, such as humans, birds, pigs and horses. Birds are a major natural reservoir of influenza A virus~\cite{sec1_14},~\cite{sec1_15}, and the virus can infect both human and pigs~\cite{sec1_16}. Pigs are also considered as an intermediate host of influenza A viruses between humans and birds~\cite{sec1_17}. Once a virus mutates through reassortment between different hosts, it can produce a life-threatening risk to human populations, as it no longer needs an intermediate host to transmit between people.

Therefore, the transmission of influenza viruses is not limited to human-to-human contact, but also includes animal-to-human (zoonosis) and animal-to-animal (enzootic) transmission~\cite{sec1_14}. Zoonotic infections can either be dead-end transmissions or lead to a pandemic in the human population after accumulating enough adaptive mutations to sustain transmissions between people, which then regularly circulates as a seasonal influenza virus~\cite{sec1_14},~\cite{sec1_18}. 

It is difficult to determine the origin of each virus during a virus outbreak because some viruses can cross species barriers. In this case, swine-origin viruses can be isolated from humans. The virus needs enough time to complete the adaptive mutation and accumulation process~\cite{sec1_14}. Therefore, earlier isolation of the original viral host may effectively control a viral outbreak or give an early warning of the risk of the virus.

Traditional methods are mostly laboratory-based, such as using hemagglutination inhibition (HI) assay to subtype virus or analyze receptor-binding. Laboratory-based methods are time-consuming and labor-intensive. In order to save manpower and time, various machine learning and deep learning algorithms have been used in viral host prediction, such as KNN~\cite{add_4}, random forests~\cite{add_4},~\cite{{add_5}}, ANN~\cite{add_3} and decision tree~\cite{add_6}. However, some previous research manually selected balanced data set~\cite{add_4},~\cite{add_2}, and some only used relative small data set~\cite{add_3} or encode the sequence as a sparse matrix~\cite{add_3},~\cite{sec1_22}. Novel deep learning techniques, such as convolutional neural networks (CNN), have also been applied in this field, but thus far only to avian and human viruses~\cite{sec1_26}.

Our study uses both classic machine learning techniques and deep learning techniques to infer the origin host of influenza viruses. Evolutionary features of viral sequences were extracted by the PSI-BLAST program and fed into four traditional machine learning models (sequence alignment-based method). Word encoding and word embedding were used for converting viral sequences to numerical vectors before feeding into a deep learning model (sequence alignment-free method). 

The paper is structured as follows: In Section 2, we introduce the data and methods used in the paper. In Section 3, we illustrate the hyperparameters and architecture of models. This is followed by the experimental results in Section 4. Finally, we provide some discussions and conclusions in Section 5. 

\section{Materials and methods}

\subsection{Data Collection}
Complete influenza A virus protein sequences isolated from avian, swine and human samples were collected from the GISAID~\cite{sec2.1_1} database (status 2020-09-25). Only hemagglutinin (HA) protein sequences are used, as HA is the most dominant protein for immunity response and helping the virus bind to target hosts~\cite{sec2.1_2}. A protein sequence usually has 20 types of amino acids, which can be represented by a one-letter symbol: A, R, N, D, C, Q, E, G, H, I, L, K, M, F, P, S, T, W, Y and V. However, sometimes a protein sequence may also include other letters to represent uncertain or unknown amino acids, such as X, B and Z~\cite{sec2.1_3}. The sequences containing such letters were removed from our data set.

\begin{figure}[h]
\setlength{\abovecaptionskip}{0.cm}
\centering
\includegraphics[width = 3.5 in]{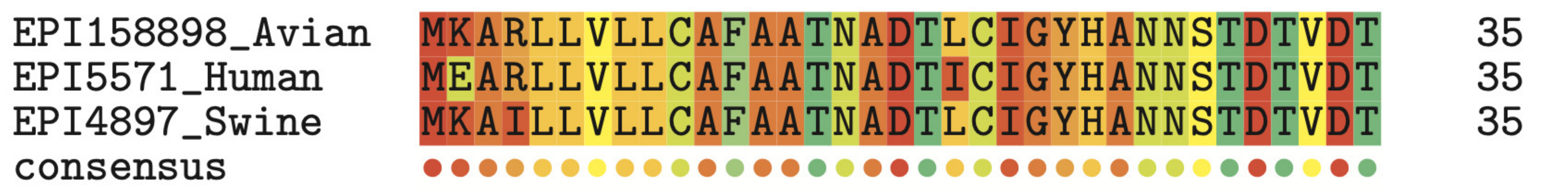}
	\caption{\small Example of highly similar sequences with different host. The strain name of each sequence is shown as follows: EPI158898\_Avian: \emph{A/chicken/AR/30402/1999}; EPI5571\_Human: \emph{A/South Carolina/1/18}; EPI4897\_Swine: \emph{A/swine/Wisconsin/1/61}. The percent identity for these sequences is above 98\%, but they were isolated in different year and from different hosts.}
\label{fig_similarity}
\end{figure}

\begin{figure}[h]
\setlength{\abovecaptionskip}{0.cm}
\centering
\includegraphics[width = 0.9\linewidth]{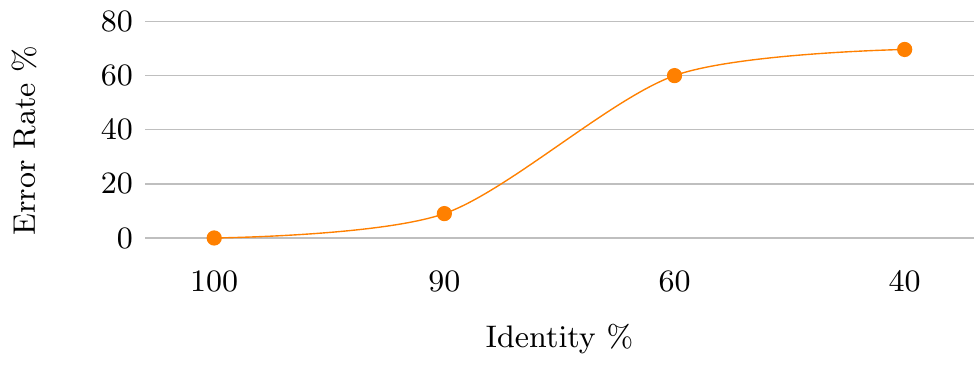}
	\caption{\small Percentage of mistakenly deleted sequences based on different identity ratio. Error rate = $\frac{\text{Number of misclassified sequences}}{\text{Total number of sequences}}$}
\label{fig_cdhit}
\end{figure}

The original data set includes redundant data, and we removed all identical sequences and produced a non-redundant data set used in this research. We did not apply any further sequence reduction steps as the influenza virus can infect across different species and may results in a high similarity between different hosts' sequences, as shown in Fig.~\ref{fig_similarity}. Therefore, simply removing similar sequences would reduce sequence diversity and mislead the results, as shown in Fig.~\ref{fig_cdhit}.

Only sequences belonging to a single host (human, avian and swine) were used in the study, which means each sequence is assigned only one label. Multi-label research is beyond the scope of this paper. Therefore, a total of 60,087 sequences have been used. Table~\ref{tab_dataset} lists the number and proportion of instances in each class.

\begin{table}[h]
\setlength{\abovecaptionskip}{0.cm}
\caption{\small Data Set}
\label{tab_dataset}
\begin{threeparttable}
\centering
\includegraphics[width = 2.5 in]{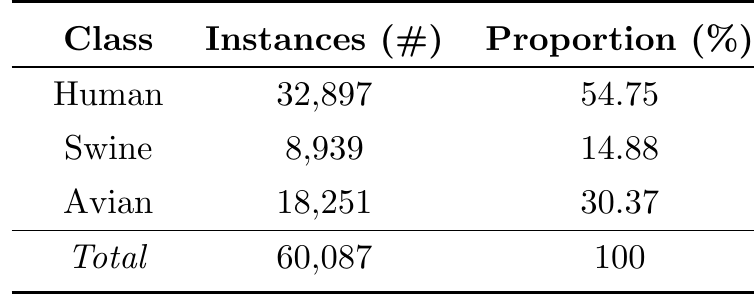}
\begin{tablenotes}
	\item\emph{Before filtering: avian (32,299),  human: (133,269), swine (15,265) and total (180,833).} 
\end{tablenotes}
\end{threeparttable}
\end{table}

\subsection{Features Derived from Position-Specific Scoring Matrix}
\subsubsection{Position-Specific Scoring Matrix}
The position-specific scoring matrix (PSSM)~\cite{sec2.2_1} is a scoring matrix that contains a highly informative representation of protein sequences and is widely used to extract evolutionary features from sequences. A PSSM can be derived by PSI-BLAST (Position-Specific Iterated BLAST) program~\cite{sec2.2_2}. 

A PSSM is an $L\times20$ matrix for a query protein sequence with $L$ length, as a protein sequence usually has 20 types of amino acids. The more intuitive way to represent the PSSM for a sequence $a_1a_2\ldots a_L$ is as follows:

\vspace{-0.3 cm}
\begin{equation}\label{equ:pssm}
\mbox{\small\( %
{\rm PSSM}_{original}=\left(\begin{matrix}&A&R&\ldots&V\\a_1&p_{1,1}&p_{1,2}&\ldots&p_{1,20}\\a_2&p_{2,1}&p_{2,2}&\ldots&p_{2,20}\\\cdots&\cdots&\cdots&\cdots&\cdots\\a_L&p_{L,1}&p_{L,2}&\ldots&p_{L,20}\\\end{matrix}\right),
\)} %
\end{equation}

Each $p_{i,j}$ is the score of amino acid $a_i$ mutated to $a_j$, it can also represent the probability of mutation by using the sigmoid function to restrict the score into the range [0, 1]:

\vspace{-0.3 cm}
\begin{equation}
p_{i,j}=\frac{1}{\left(1+e^{-p_{i,j}}\right)},i=1,2,\ldots,L;j=1,2,\ldots20 ,
\end{equation}

We apply a standalone version of PSI-BLAST~\cite{sec2.2_4} developed by the NCBI to run PSI-BLAST iteratively. The database used for searching is a non-redundant (NR) database, and the parameters of the PSI-BLAST program are set to their default values ($E$-value = 0.001, number of iterations = 3). 

The original PSSMs are variable in size and thus cannot be fed directly into many machine learning models. Therefore, we propose PSSM-based sequence encoding schemes to overcome this hindrance. We first introduce a residue grouping rule to reduce the complexity of proteins and reduce unnecessary computations.

\subsubsection{Residue Grouping Rule}
20 kinds of amino acids can be grouped into 10 types as they have similar functional or structural characteristics in proteins~\cite{sec2.3_2}: G1 (F, \textbf{Y}, W), G2 (M, \textbf{L}), G3 (I, \textbf{V}), G4 (A, T, \textbf{S}), G5 (\textbf{N}, H), G6 (Q, \textbf{E}, D), G7 (R, \textbf{K}), G8 (C), G9 (G) and G10 (P).
By implementing residue grouping rules to each column of original PSSM, a grouped-PSSM (GPSSM) with $L\ \times\ 10$ dimension can be formed:

\vspace{-0.3 cm}
\begin{equation}\label{equ:gpssm}
\mbox{\small\( %
{\rm PSSM}_G=\left(\begin{matrix}&G_1&G_2&\ldots&G_{10}\\a_1&g_{1,1}&g_{1,2}&\ldots&g_{1,10}\\a_2&g_{2,1}&g_{2,2}&\ldots&g_{2,10}\\\cdots&\cdots&\cdots&\cdots&\cdots\\a_L&g_{L,1}&g_{L,2}&\ldots&g_{L,10}\\\end{matrix}\right),
\)} %
\end{equation}

where 

\vspace{-0.3 cm}
\begin{equation}
g_{i,j}=\frac{\sum p_{i,G_j}}{\left|G_j\right|},
\end{equation}

The GPSSM is produced based on the original PSSM~\eqref{equ:pssm}, thence $\sum p_{i,G_j}$ means the score of amino acid $a_i$ mutated to an amino acid that belongs to group $j$. $L$ is the length of sequences; $i = 1, 2, \ldots, L$; $\left|G_j\right|$ is the number of amino acids types in group $j$. For instances, if $i=j=1$, then $\left|G_1\right|=3$, $g_{1,1}=\nicefrac{(p_{1,F}+\ p_{1,Y}+\ p_{1,W})}{3}$

The following proposed feature sets (EG-PSSM, GDPC-PSSM and ER-PSSM) are derived from GPSSM~\eqref{equ:gpssm}.

\subsubsection{EG-PSSM}
Because the length of the input sequences can vary, so can original PSSMs~\eqref{equ:pssm} and GPSSMs~\eqref{equ:gpssm}. As a result, they cannot be directly used in many machine learning models. One intuitive and simple way to overcome this problem is to also apply the residue grouping rule to each row of GPSSM~\eqref{equ:gpssm}. Therefore, each sequence can produce a $10\times10$ matrix. Reformatting the matrix in rows from top to bottom and left to right, a $1\times100$ feature vector is extracted from a GPSSM~\eqref{equ:gpssm}.

\vspace{-0.3 cm}
\begin{equation}
{\rm PSSM}_{EG}=\left(\begin{matrix}E_{G_1,G_1}&E_{G_1,G_2}&\begin{matrix}\cdots&E_{G_{10},G_{10}}\\\end{matrix}\\\end{matrix}\right)^T,
\end{equation}

where 

\vspace{-0.3 cm}
\begin{equation}
E_{G_i,G_j}=\frac{\sum g_{G_i,G_j}}{\left|G_i\right|}, i,j = 1 \ldots 10,
\end{equation}

\subsubsection{GDPC-PSSM}
The traditional dipeptide composition (DPC)~\cite{sec2.3_3} captures the composition information of amino acids and partial local-order information in protein sequences. Original DPC acts directly on raw sequence data and gives a 400-dimensional feature vector for each sequence, but it can be further extended to PSSM~\cite{sec2.3_4}. Therefore, each $L\ \times\ 10$ GPSSM~\eqref{equ:gpssm} can be further defined as 100-dimensional feature vector by grouped dipeptide composition encoding:

\vspace{-0.3 cm}
\begin{equation}
{\rm PSSM}_{GDPC}=\left(\begin{matrix}D_{1,1}&D_{1,2}&\begin{matrix}\cdots&D_{10,10}\\\end{matrix}\\\end{matrix}\right)^T,
\end{equation}

where 

\vspace{-0.3 cm}
\begin{equation}
D_{i,j}=\frac{1}{L-1}\sum_{k=1}^{L-1}{g_{k,i}\times g_{k+1,j}}\ i,j=1,2,\ldots,10,
\end{equation}

Each $g_{k,i}$ is the value of row $k$ and column $i$ in the GPSSM~\eqref{equ:gpssm}.

\subsubsection{ER-PSSM}
The third proposed sequence representation is adapted from RPSSM~\cite{sec2.3_5}, which computes the pseudo-composition of dipeptide in sequences. Same as GDPC-PSSM, RPSSM also extracts the partial local sequence order information in sequences. Original RPSSMs only compute the pseudo-composition of any two adjacent amino acids. We extend the computation of RPSSM for any two amino acids $a_ka_{k+t}$ with gap $t$ in sequences and extract a $1\times910$ feature vector per sequence: 

\vspace{-0.3 cm}
\begin{equation}
\resizebox{.9\hsize}{!}{${\rm PSSM}_{ER}=\left(\begin{matrix}M_{1,1,1}&M_{1,2,1}&\begin{matrix}\cdots&\begin{matrix}M_{10,10,9}&\begin{matrix}T_1&\begin{matrix}\cdots&T_{10}\\\end{matrix}\\\end{matrix}\\\end{matrix}\ \\\end{matrix}\\\end{matrix}\right)^T $},
\end{equation}

where

\vspace{-0.3 cm}
\begin{equation}
\begin{split}
M_{i,j,t}=\frac{1}{L-t}\sum_{k=1}^{L-t}\frac{\left(g_{k,i}-g_{k+t,j}\right)^2}{2},\ \\
i,j=1,2,\ldots,10;t=1,2,\ldots,9
\end{split}
\end{equation}

and 

\begin{equation}
T_i=\frac{1}{L}\sum_{k=1}^{L}{\left(g_{k,i}-\bar{G_i}\right)^2,\ \ \ }i,j=1,2,\ldots,10.
\end{equation}

$\bar{G_i}$ is the average of values of GPSSM~\eqref{equ:gpssm} in column $i$, $T_i$ computes the average pseudo-composition of all the amino acids in the protein sequence corresponding to column $i$ in GPSSM~\eqref{equ:gpssm}.

\subsection{Features Learned from N-grams}

Traditional machine learning methods require manual data preprocessing and feature extraction to extract representative features of each protein sequence. The feature extraction process requires prior knowledge to select suitable features. In contrast, deep learning methods can directly learn the implicit representation of protein sequences. This subsection introduces two vectorization schemes, word encoding and word embedding, to map protein sequences into numerical vectors. 

\subsubsection{Overlapping N-grams}
\begin{figure}[h]
\setlength{\abovecaptionskip}{0.cm}
\centering
\includegraphics[scale = 0.95]{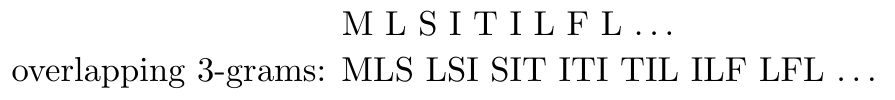}
\caption{\small Example of Overlapping Trigrams.}
\label{fig_3gram_pssm}
\end{figure}

\begin{figure}[h]
\setlength{\abovecaptionskip}{0.cm}
\centering
\subfloat[3-grams]{%
  \includegraphics[width=0.16\textwidth]{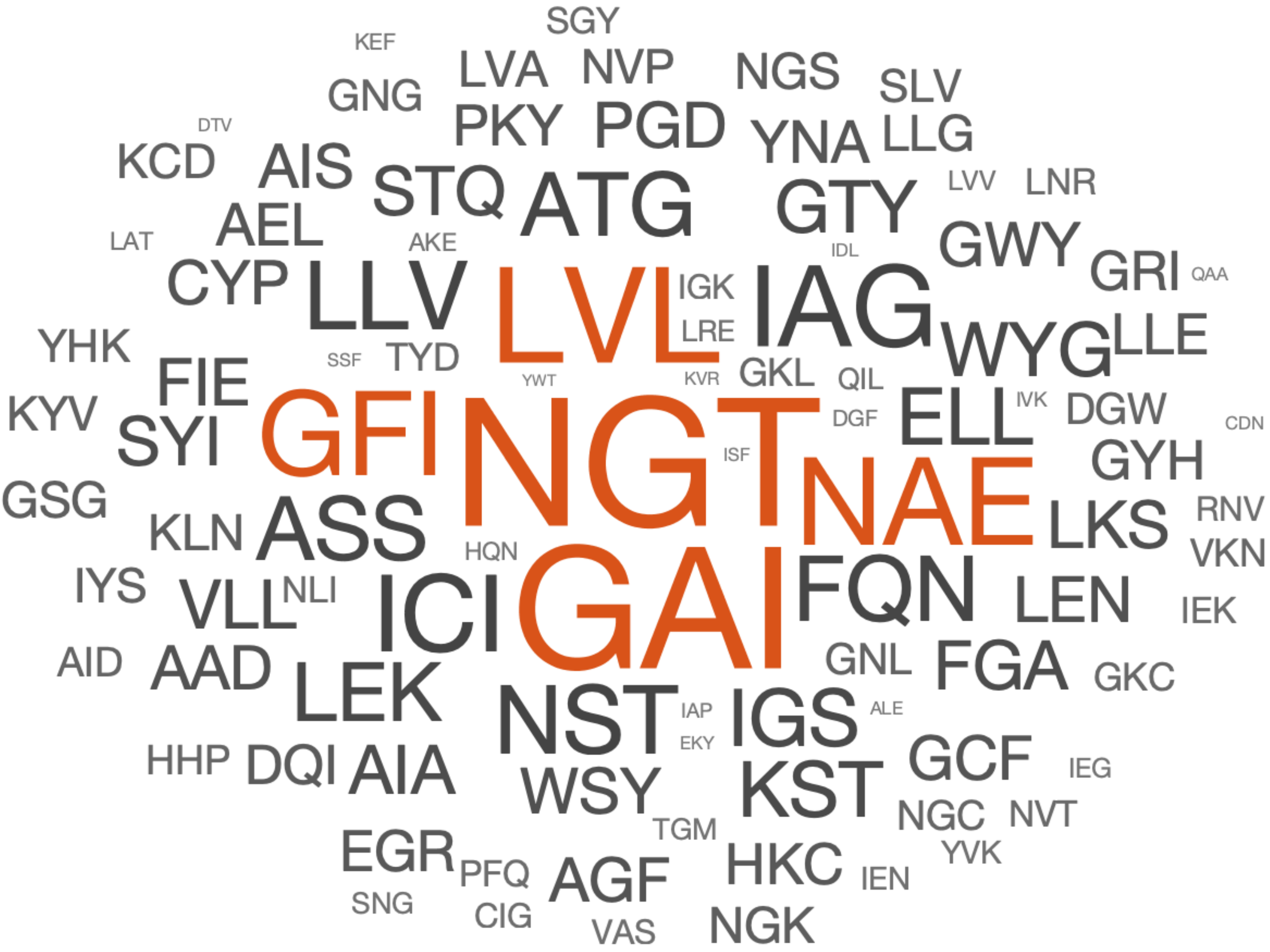}
}
\subfloat[4-grams]{%
  \includegraphics[width=0.16\textwidth]{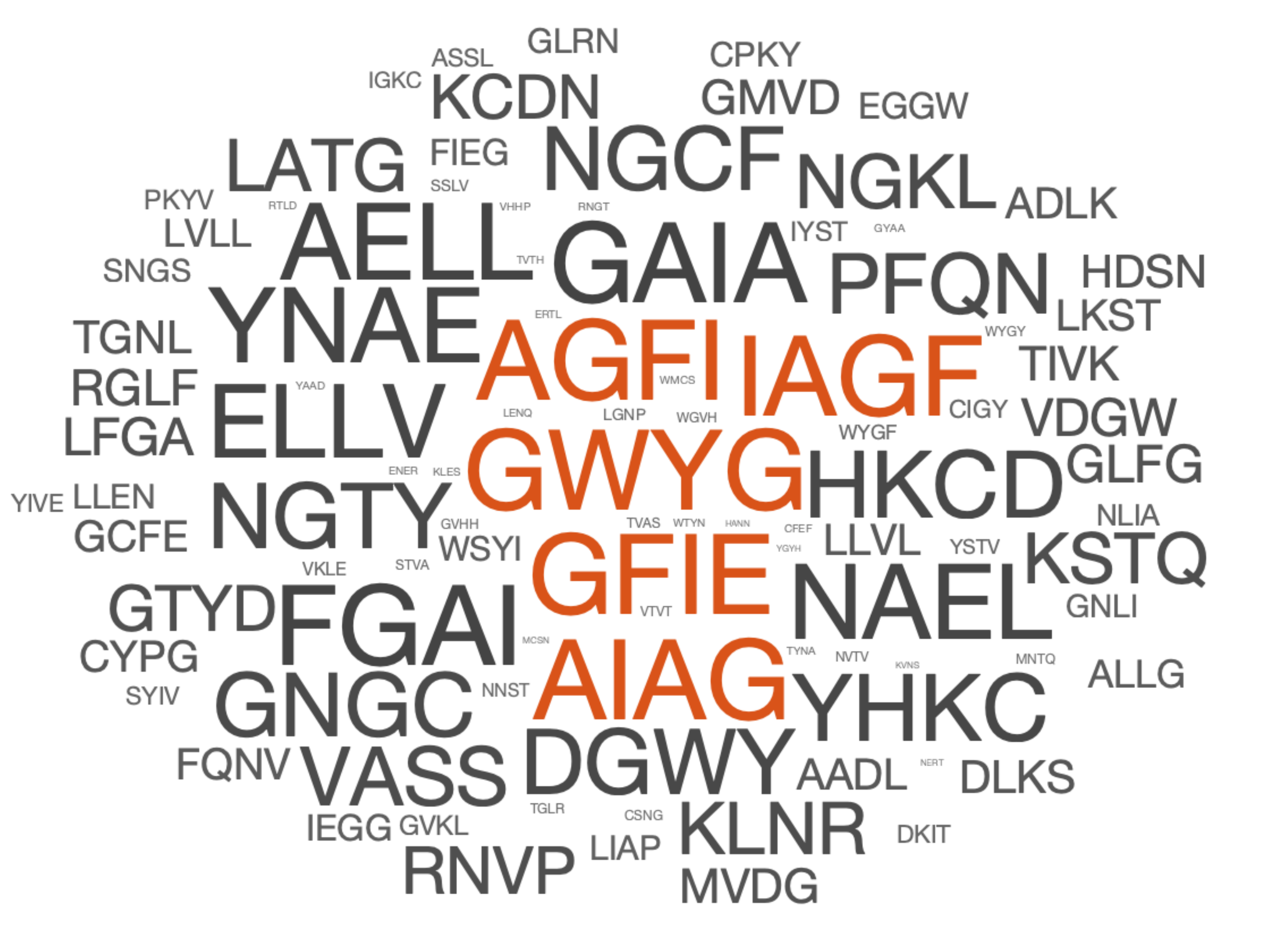}
}
\subfloat[5-grams]{%
  \includegraphics[width=0.16\textwidth]{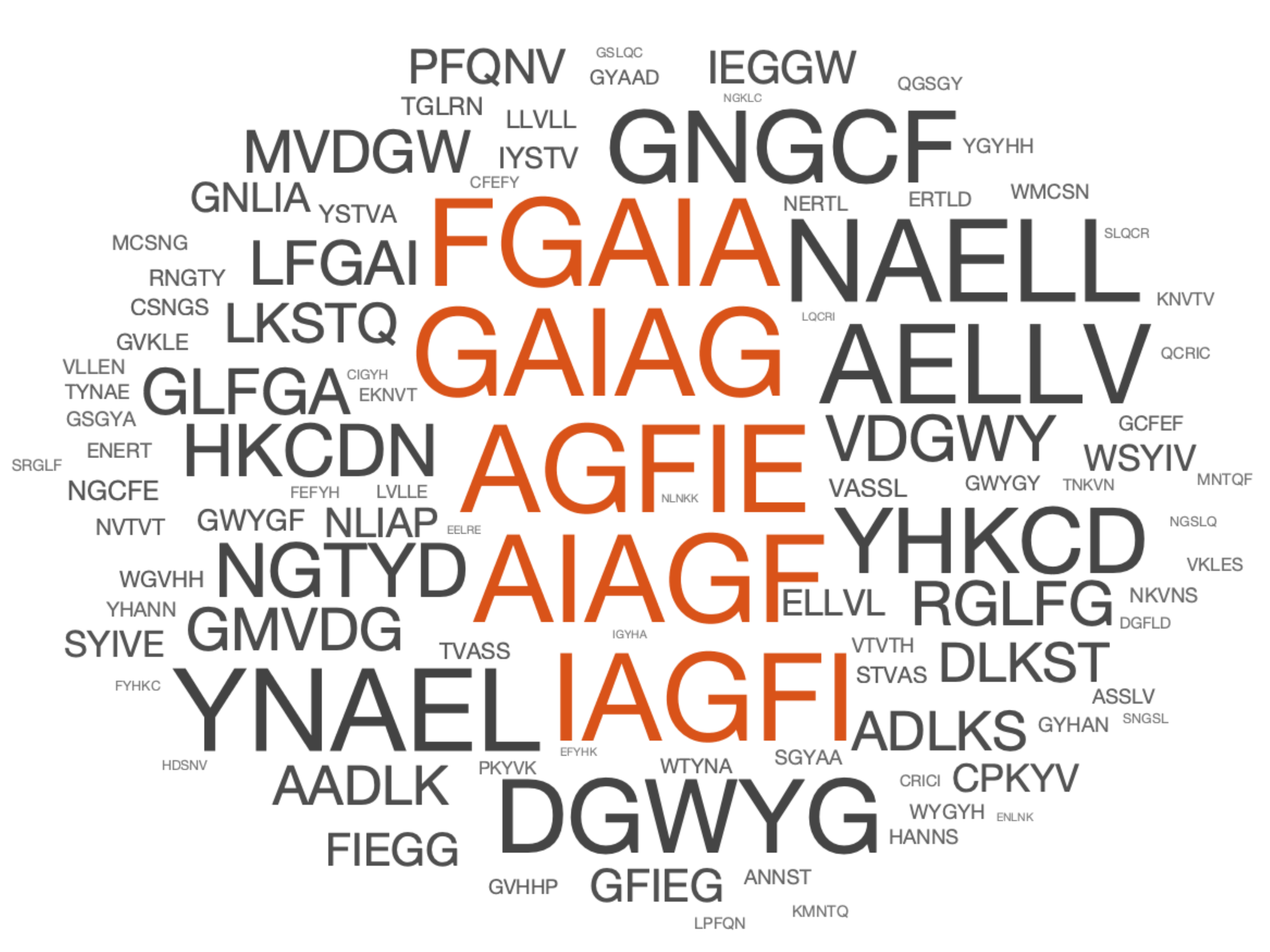}
}
\caption{\small Word Clouds of N-grams}
\label{fig_wordcloud}
\end{figure}

A protein sequence is morphologically similar to a text sentence, except that the text is composed of words but the protein sequence is formed by amino acid letters. Therefore, we split the sequence into overlapping n-grams (n is ranging from 3 to 5) to transform a protein sequence into a protein "sentence" of n-grams. An n-gram is a protein "word" with successive n amino acids. Fig.~\ref{fig_3gram_pssm} is an example of overlapping 3-grams for a protein sequence, and Fig.~\ref{fig_wordcloud} shows the word clouds of n-grams for all sequences.

\subsubsection{Word Encoding}
\begin{figure}[h]
\setlength{\abovecaptionskip}{0.cm}
\centering
\includegraphics[scale = 0.8]{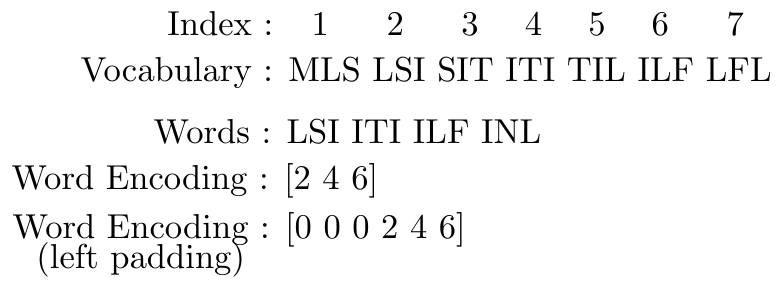}
\caption{\small Example of Word Encoding.}
\label{fig_wordEnc}
\end{figure}

Word encoding, also called indexed-based encoding, maps words to numbers, as shown in Fig.~\ref{fig_wordEnc}. Compared with commonly used one-hot encoding, word encoding can generate non-sparse vectors and is more efficient. However, word encoding is not trainable and unexplainable for models that need to learn feature weights as it loses the relationship between words.

\subsubsection{Word Embedding}

\begin{figure}[h]
\setlength{\abovecaptionskip}{0.cm}
\centering
\subfloat[3-grams]{%
  \includegraphics[width=0.16\textwidth]{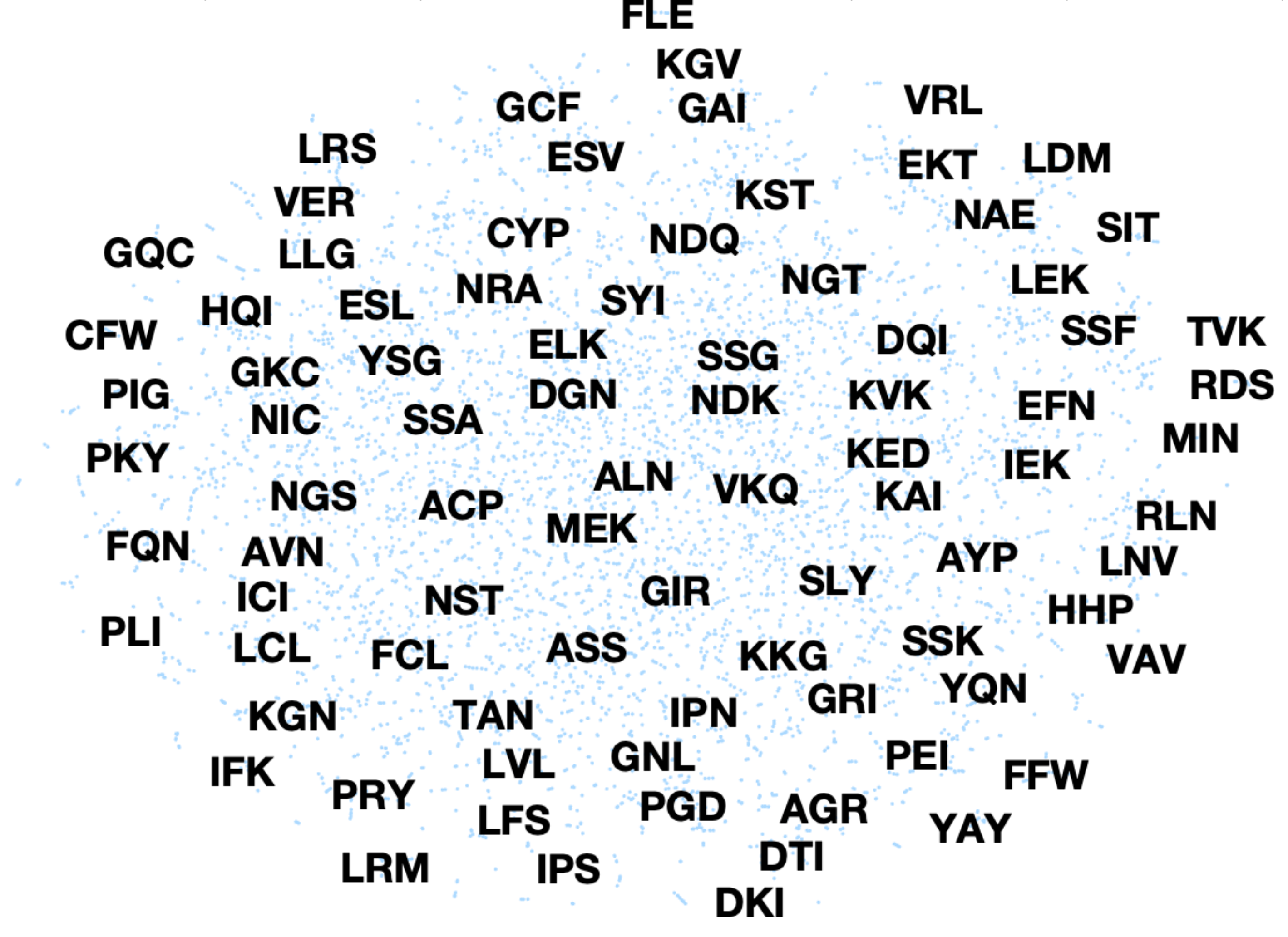}%
}
\subfloat[4-grams]{%
  \includegraphics[width=0.16\textwidth]{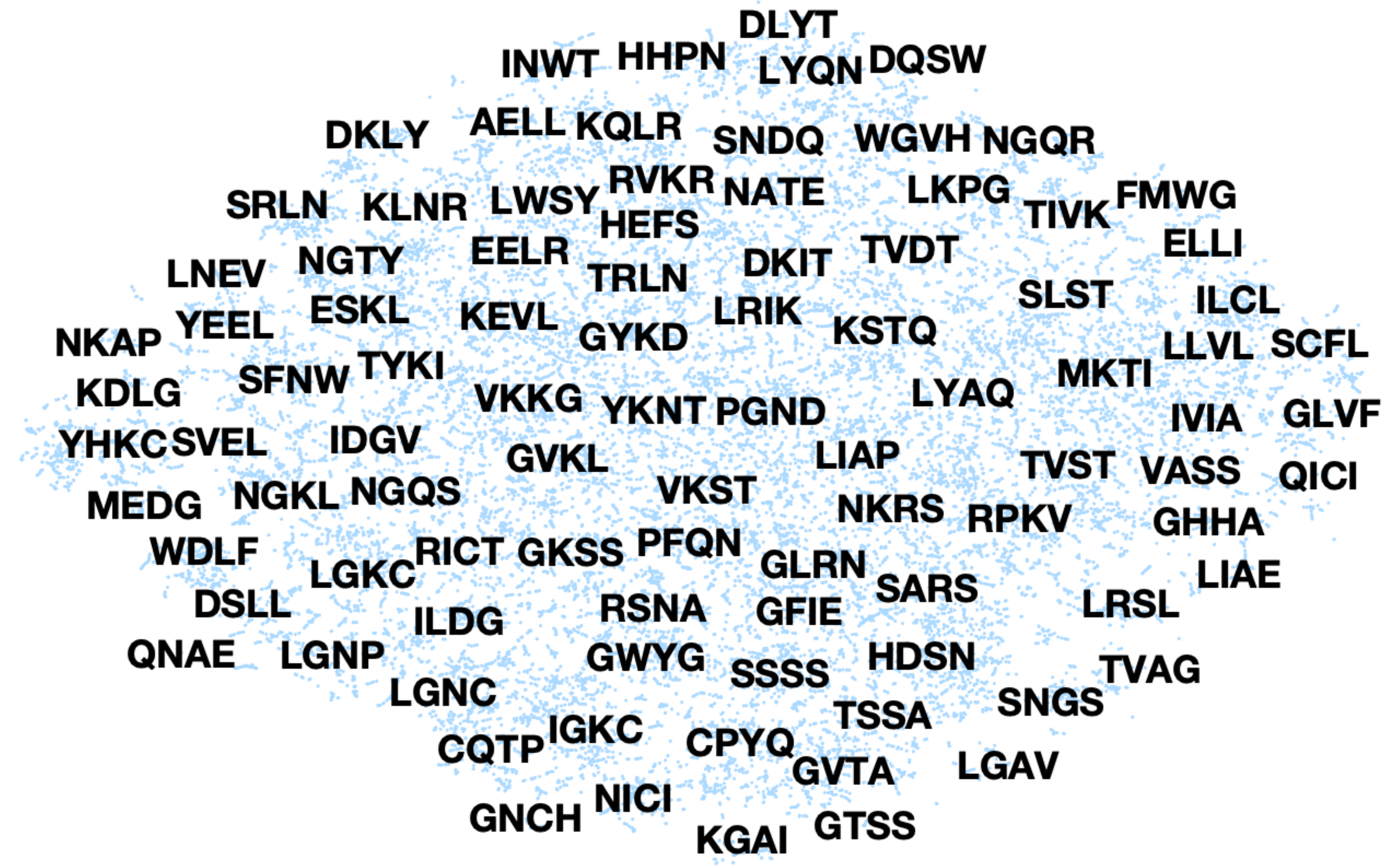}%
}
\subfloat[5-grams]{%
  \includegraphics[width=0.16\textwidth]{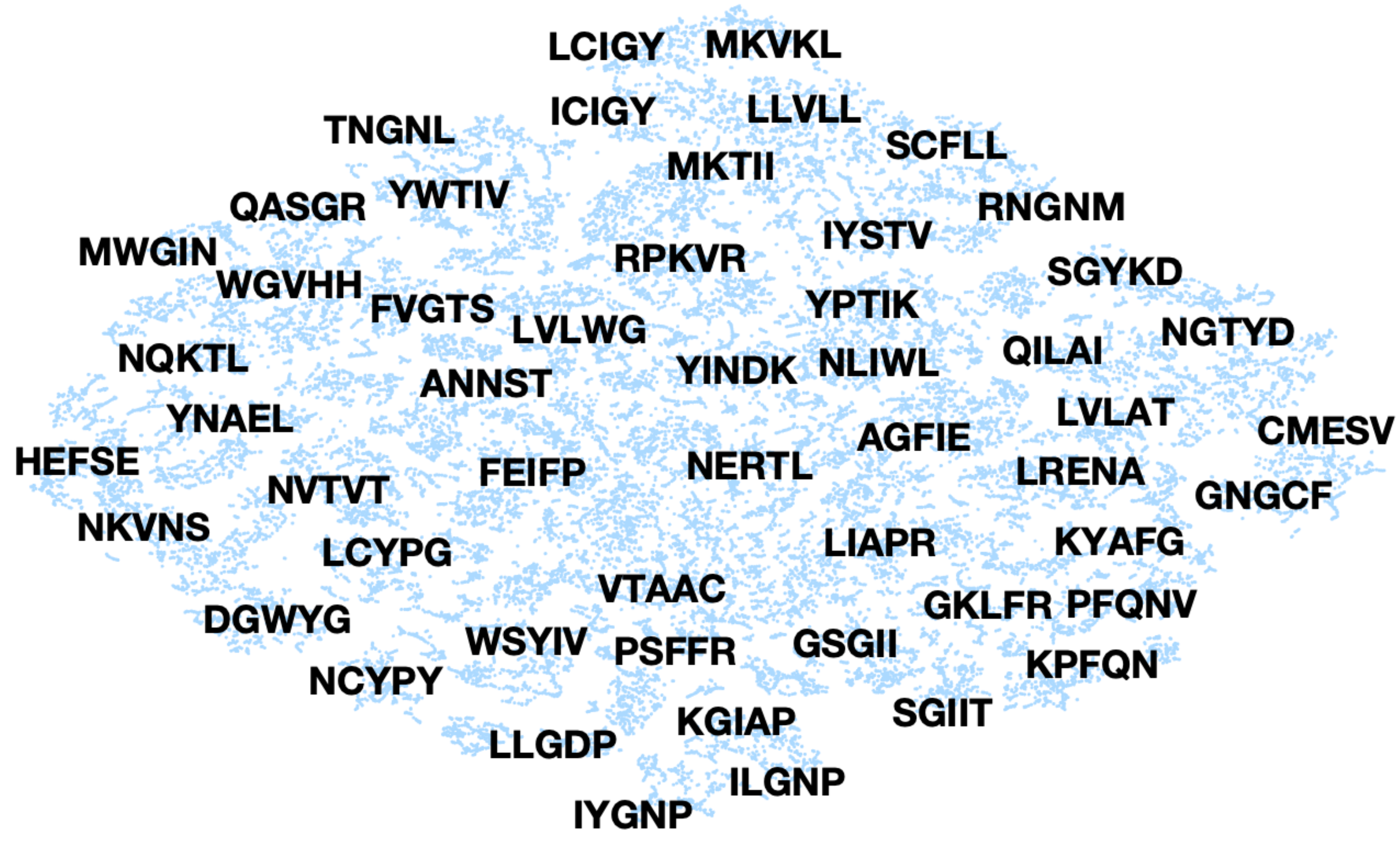}%
}
\caption{\small Word Embeddings Visualization Using 2-D t-SNE}
\label{fig_emb}
\end{figure}

Word embedding compensates for the drawbacks of word encoding and one-hot encoding. It cannot only produce dense vectors but also capture the relationship between similar words. Popular implementations of word embedding include Word2Vec~\cite{sec1_21}, but it lacks domain-specific words. Therefore, we generated a custom word embedding from the dataset only used for training and mapped the n-grams of each sequence to the embedding vectors. Each n-gram is represented as a vector of size $N$, and a protein sequence is represented as a $L \times N$, where $L$ is the length of the sequence (number of n-grams in the sequence) and $N$ is the embedding dimension. Fig.~\ref{fig_emb} depicts the visualization of word embedding using 2-d t-SNE.

We left-pad and truncate the sequence with the most frequent sequence length to unify the dimensionality of matrices, which means that most sequence information will be retained, but more noise will be introduced to the shortest sequence, and the longest sequence information will be discarded.

\subsection{Machine Learning Techniques}
\subsubsection{RUSBoosted Tree}
Data sampling with boosting algorithms are applied to tackling the problem of class imbalance. Commonly used data samples methods include oversampling (enriching the minority class) and undersampling (decreasing the majority class). Random undersampling boosting (RUSBoost) algorithm~\cite{sec2.4_3}, as its name implies, combines undersampling methods with boosting algorithms. Compared with other oversampling methods, e.g., SMOTEBoost, RUSBoost is computationally cheaper and more efficient.

\subsubsection{Extreme Gradient Boosting}
Extreme Gradient Boosting (XGBoost) ~\cite{sec2.4_7} is a highly efficient and scalable implementation of gradient boosting algorithms. Gradient boosting algorithms are similar to AdaBoost but use gradient descent to optimize the derivable loss function when adding the new models. XGBoost can handle stubborn issues in the data science area, such as it can solve missing values and sparse data in an automatic way. One of the biggest advantages of XGBoost is that it provides parallel training to speed up the training process and can handle large datasets. 

\subsubsection{Random Forest} 
Decision trees have higher variance and lower bias, while the bagging algorithm aims to reduce the variance of the model. Therefore, the combination of bagging and decision tree (random forest) improves the overall performance of the model. Random forest~\cite{sec2.4_4} considers only a small part of all features in each split. It introduces more randomness to each decision tree. In contrast to boosting-based ensembles, bag-based ensembles are prone to construct deep trees, which means bag-based ensembles are more complex than boosting-based ensembles. Therefore, bag-based ensembles may require more training time than boosting-based ensembles but leave out the validation process to estimate generalization performance. 

\subsubsection{Support Vector Machine}
Support vector machine (SVM) is one of the best "off-the-shelf" supervised learning algorithms~\cite{sec2.4_5}. SVM can not only classify linearly separable data but also can classify non-linearly separable data by introducing a kernel trick.  Kernel tricks help SVMs handle high-dimensional data (even infinite-dimensional data) well by mapping lower-dimensional data into higher dimensional data but without explicitly transforming them. We use the Gaussian kernel as kernel function and the one-vs-all strategy to construct the multi-class SVM.

\subsubsection{Convolutional Neural Network}
Convolutional neural networks (CNN or ConvNets) often appear in areas related to computer vision, such as facial recognition, object recognition, and autonomous vehicles. CNN initially took images as input and expanded to include non-image data such as time series, text, and audio data. Contrary to traditional machine learning, CNN learns features of the data in each hidden layers. A CNN often includes three hidden layers: the convolutional layer for learning certain features, the activation layer for activating features, and the pooling layer for reducing the number of network parameters.

\subsection{Model Parameters and Implementation}

Four classic machine learning models (SVM, RF, RUSBoost and XGBoost) were evaluated using optimized parameters (see Table~\ref{tab_tunning} for the tunning range). We used Bayesian Optimization to automatically adjust hyperparameters in 30 iterations and at most 40,000 seconds. The deep learning model (CNN) was evaluated using fixed parameters (see Table~\ref{tab_cnn_params} for details).

RF, RUSBoost, SVM and CNN were implemented in MATLAB 2020a, and XGBoost was implemented by XGBoost Python Package. 

\begin{table}[h]
\setlength{\abovecaptionskip}{0.cm}
\centering
\caption{\small Hyperparameter Setting for Classic ML Models}
\includegraphics[width = \linewidth]{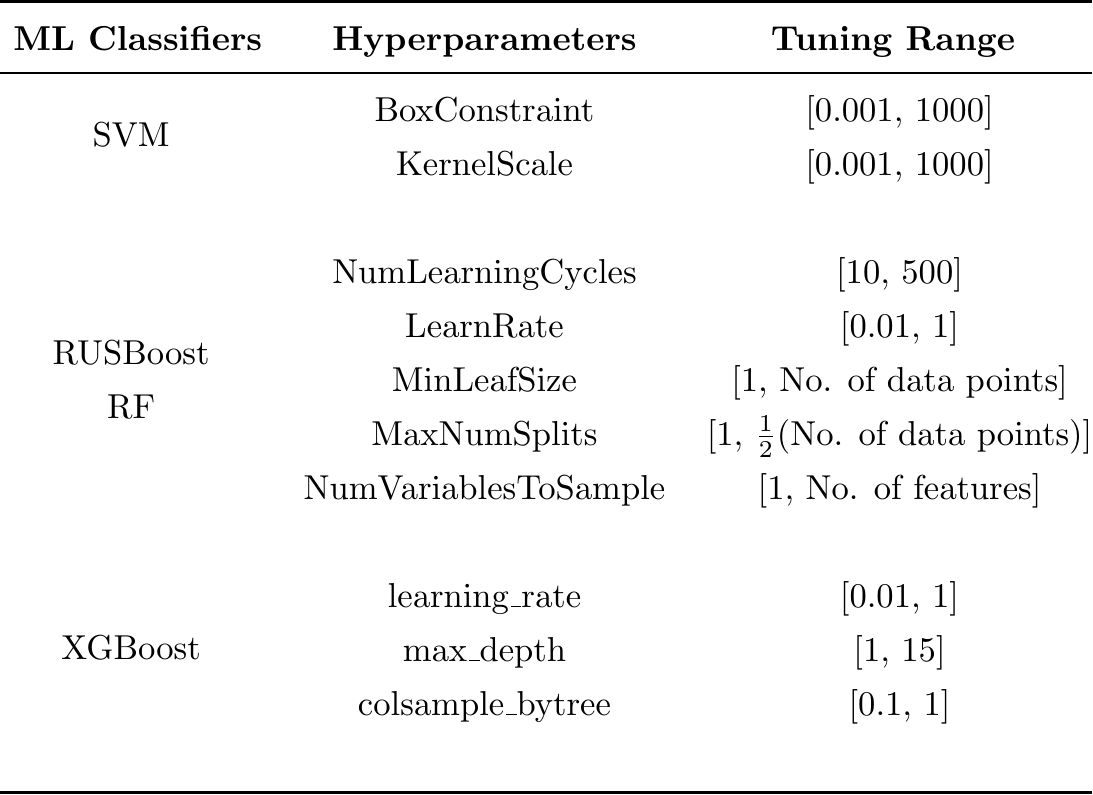}
\label{tab_tunning}
\end{table}

\begin{table}[h]
\setlength{\abovecaptionskip}{0.cm}
\centering
\caption{\small Hyperparameter Setting for CNN}
\includegraphics[width = 2 in]{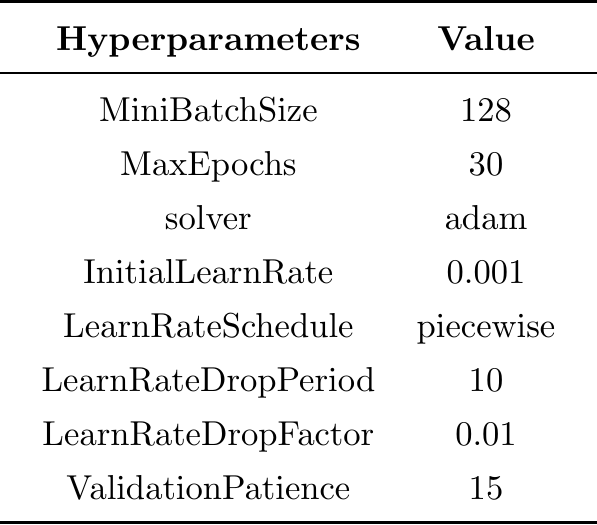}
\label{tab_cnn_params}
\end{table}

\subsection{CNN Architecture}
\begin{figure}[h]
\setlength{\abovecaptionskip}{0.cm}
\centering
\includegraphics[width = \linewidth]{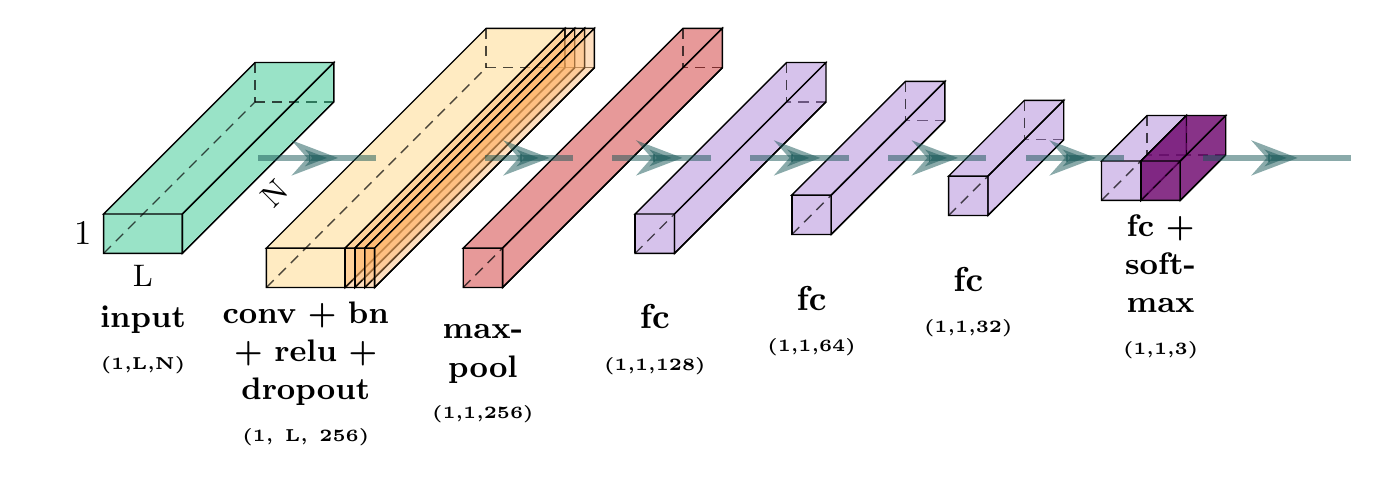}
\caption{\small Example of the CNN Architecture}
\label{fig_cnn_architecture}
\end{figure}

We designed a simple CNN for classifying protein n-grams, as shown in Fig.~\ref{fig_cnn_architecture}. The CNN used in this work contains one input layer (\emph{input}), one convolution layer (\emph{conv}), one batch normalization layer (\emph{bn}), one ReLU activation layer (\emph{relu}), one dropout layer (\emph{dropout}), one max pooling layer (\emph{max-pool}), four fully connected layers (\emph{fc}) and one softmax layer (\emph{softmax}). Each protein feature matrix has $(L, N)$ dimension, where $L$ is the length of the protein sentences after padding or cropping, and $N$ is the embedding dimension of that sequence. Typically, CNN takes the image input with size $(height, width, color \ channels)$. When it is applied to text classification, the height of the input is set as 1. Thence, the output size of the the convolution layer (\emph{conv}) is $(1, L, 256)$. The convolution layer (\emph{conv}) has 256 filters of size $[1 \ ngram]$, where $ngram$ is the size of protein words. Two parameters of the convolution layer (\emph{conv}) and max pooling layer (\emph{max-pool}), which are not mentioned in Table~\ref{tab_cnn_params}, are step size (stride) and padding value that were set as $[1 \ 1]$ and 0, respectively. The dropout rate in the dropout layer (\emph{dropout}) is 0.2. 

\section{Model Evaluation}
\subsection{Cross-validation}

\begin{figure}[h]
\setlength{\abovecaptionskip}{0.cm}
\centering
\includegraphics[width = \linewidth]{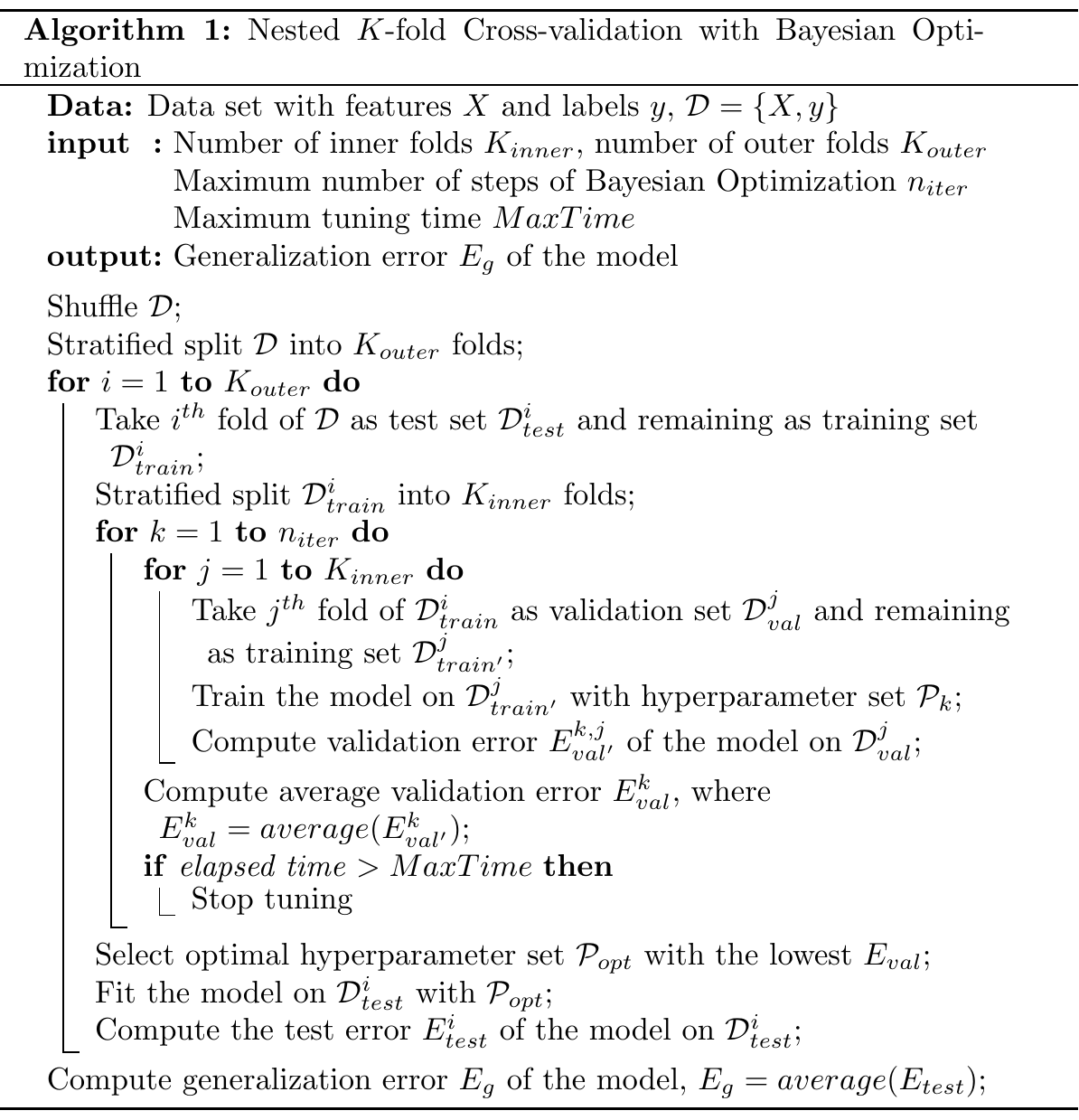}
\caption{\small Pseudo Code of Stratified Nested $K$-fold CV}
\label{fig_nestedCV}
\end{figure}

We used stratified $K$-fold cross-validation (CV) to evaluate models. The class ratio of the training set was almost the same as that of the test set. The generalization performance of models was only evaluated on the test set (unseen data to the model). Nested $K$-fold CV adds the outer $K$-fold CV for final evaluation to reduce bias when it comes to hyperparameters optimization and model selection~\cite{add_1}. Therefore, nested $K$-fold CV will take advantage of the full diversity of the data set and ensure that all data will be tested. The pseudo-code of stratified nested $K$-fold CV is shown in Fig.~\ref{fig_nestedCV}.

In this study, we chose $k_{outer}=6$ and $k_{inner}=5$. Therefore, approximately 68\%, 16\%, and 16\% of the data were used for training, validation, and testing, respectively. The detailed information on the training set and test set is shown in Table~\ref{tab_trainset}.

\begin{table}[h]
\setlength{\abovecaptionskip}{0.cm}
\centering
\caption{Training Set and Test Set}
\label{tab_trainset}
\begin{threeparttable}
\includegraphics[width = \linewidth]{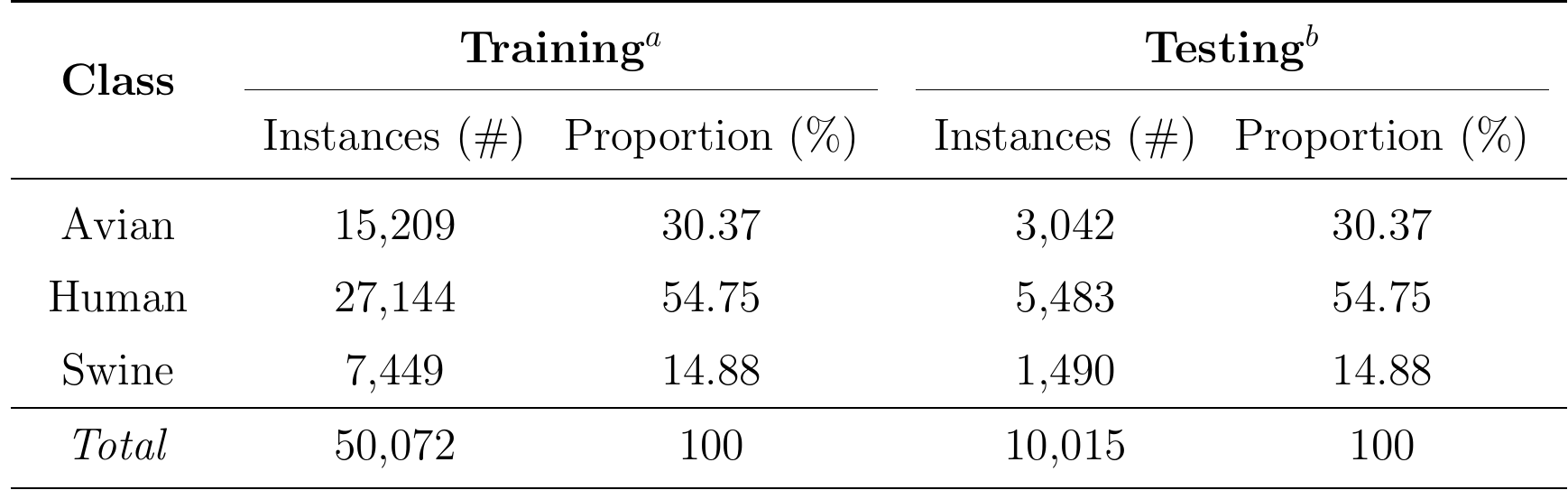}
\begin{tablenotes}
  \item[a] \emph{The training set contains 84\% of all data, 80\% of the training set used for training and 20\% of the training set used for validation (5-fold cross-validation).}     
  \item[b] \emph{The test set contains 16\% of all data and is only used in final evaluation.}
\end{tablenotes}
\end{threeparttable}
\end{table}

\subsection{Evaluation Metrics}

Evaluation measurements used in the study include \emph{$F_1$}-score and Matthews’s correlation coefficient (\emph{MCC}). The equations of measurements for each class are defined as follows:

\vspace{-0.2 cm}
\begin{equation}
F_{1_i}=\ 2\ \cdot\frac{\rm Precision_i \cdot \rm Sensitivity_i}{\rm Precision_i+ \rm Sensitivity_i},
\end{equation}

\vspace{-0.2 cm}
\begin{equation}
\resizebox{.9\hsize}{!}{${MCC}_i=\frac{TP_i\times T N_i-FP_i\times F N_i}{\sqrt{\left(TP_i+FP_i\right)\left(TP_i+FN_i\right)\left(TN_i+FP_i\right)\left(TN_i+FN_i\right)}}$},
\end{equation}

where $i=1,2,\ldots,\ N$, $N$ is the number of classes; ${\rm Sensitivity}_i=\nicefrac{TP_i}{(TP_i+FN_i)}$ and ${\rm Precision}_i=\nicefrac{TP_i}{(TP_i+FP_i)}$. $TP$ (True Positive) and $TN$ (True Negative) represent the number of data correctly predicted, $FP$ is the number of negative data misclassified as positive, and $FN$ counts the number of positive data incorrectly predicted as negative. 

For multi-class classification, the one-vs-all strategy is applied to produce \emph{$F_1$}-score for each class. The overall \emph{$F_1$}-score and overall \emph{MCC} are defined as follows:

\vspace{-0.2 cm}
\begin{equation}
{\rm Overall} \ F_{1}= \frac{\sum_{i=1}^{N}{F_{1_i}}}{\left|C\right|},
\end{equation}

\vspace{-0.2 cm}
\begin{equation}
{\displaystyle {{\rm Overall} \  MCC}={\frac {c \cdot s- {\sum_{i}^{N}{p_i}}\cdot{{t_i}}}{{\sqrt {s^{2}-{\sum_{i}^{N}{p_i^2}} }}\cdot{\sqrt {s^{2}-{\sum_{i}^{N}{t_i^2}}}}}}}
\end{equation}

where $t_{i}$ is the number of times that class $i$ truly occurred, $p_{i}$ is the number of times class $i$ was predicted, $c$ is the total number of correctly predicted data, $s$ is the total number of data items.

\section{Experimental Results}
\subsection{Sequence Alignment-based Methods}

\begin{figure}[h]
\setlength{\abovecaptionskip}{0.cm}
\centering
\includegraphics[width = \linewidth]{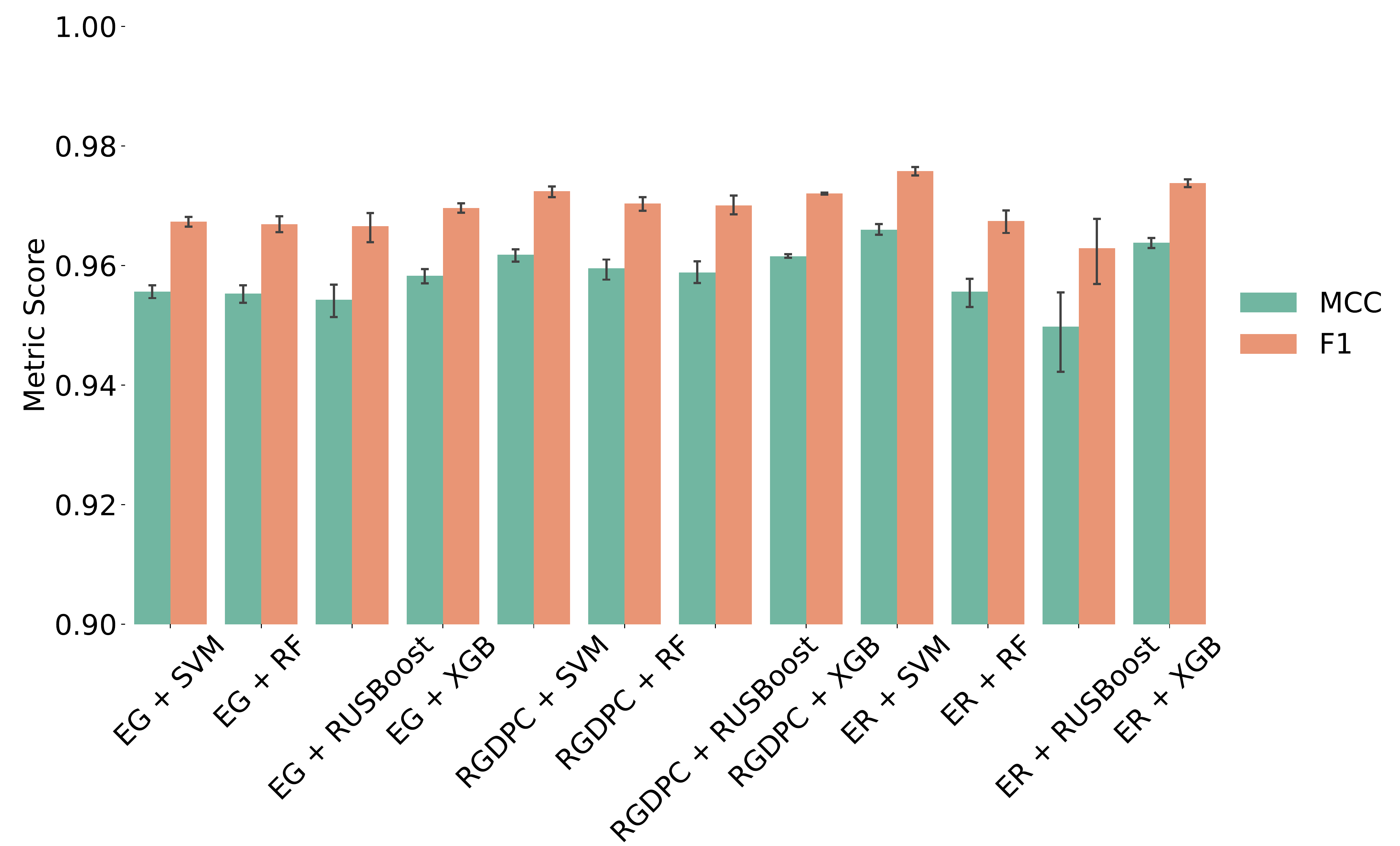}
\caption{\small Overall Performance of PSSM-based Models}
\label{fig_results_pssm_overall}
\end{figure}

Fig.~\ref{fig_results_pssm_overall} shows the overall performance of the optimized PSSM-based models on the test set. After hyperparameter optimization, the performance of all models is outstanding, but RUSBoost with ER-PSSM has highest variation.

\begin{figure}[h]
\setlength{\abovecaptionskip}{0.cm}
\centering
\includegraphics[width = \linewidth]{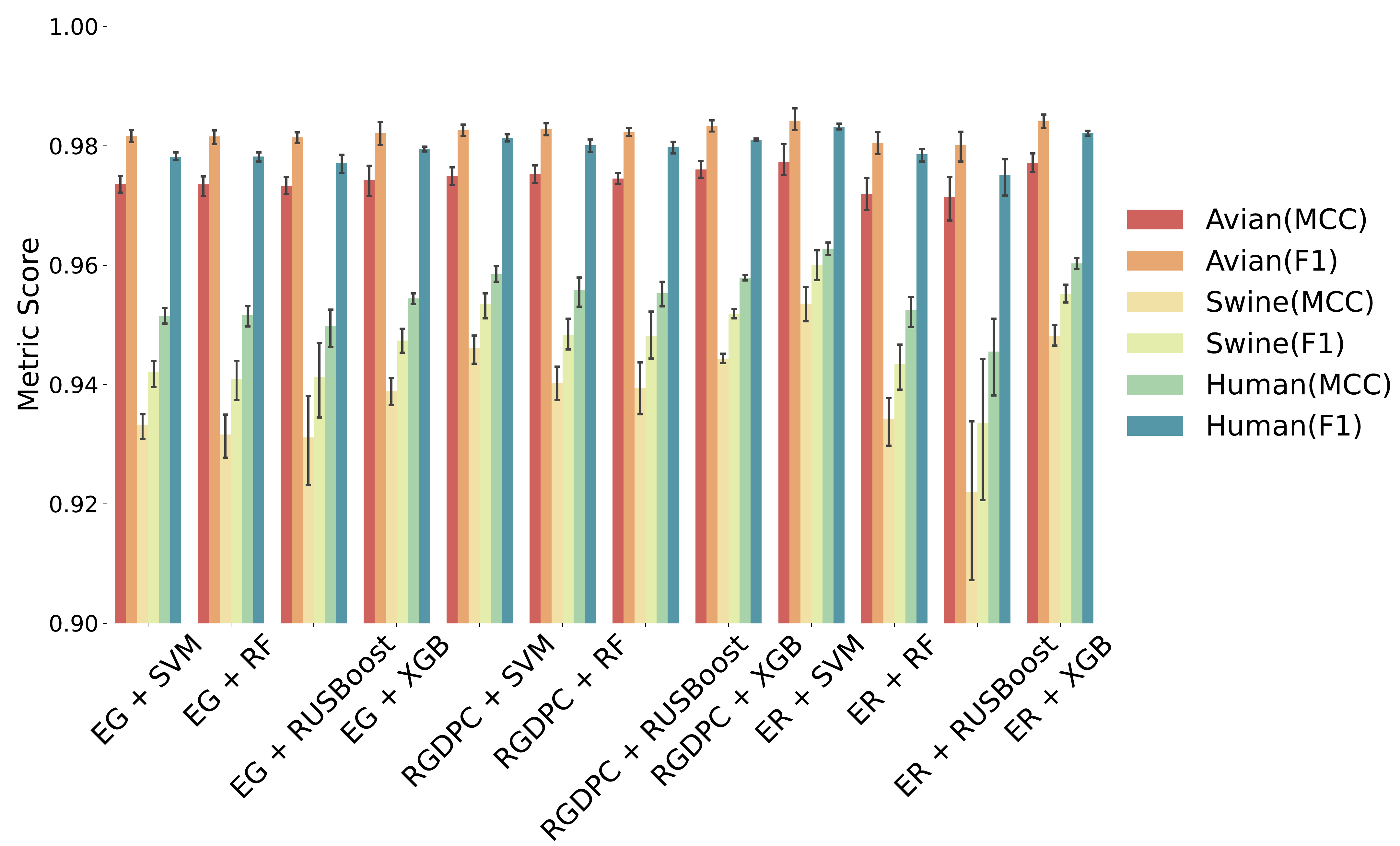}
\caption{\small Performance of PSSM-based Models in Individual Classes}
\label{fig_results_pssm_classes}
\end{figure}

The classes in the data set are unbalanced: more than half of the data belongs to human viruses, whereas only around 15\% belongs to swine viruses. The class imbalance problem means that the classifier may ignore the minority class, resulting in performance degradation for the minority class. All models yield the worst performance in swine viruses compared with human and avian (see details in Fig.~\ref{fig_results_pssm_classes}). Among all PSSM-based models, the SVM with ER-PSSM performs best in individual classes.

\subsection{Sequence Alignment-free Methods}

\begin{figure}[h]
\setlength{\abovecaptionskip}{0.cm}
\centering
\includegraphics[width = \linewidth]{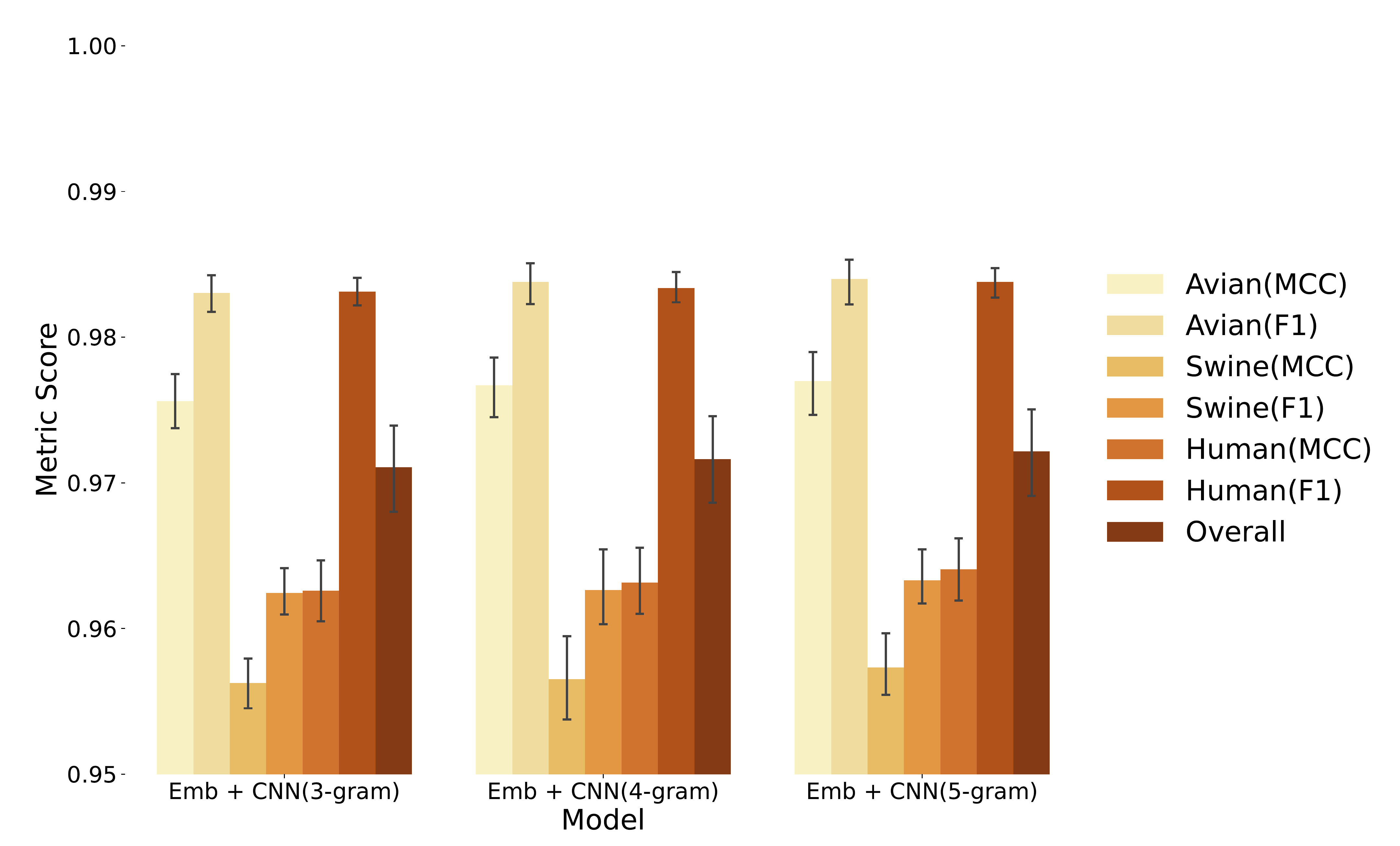}
\caption{\small Performance of Embedding-CNN in Individual Classes}
\label{fig_results_cnn_overall}
\end{figure}

\begin{figure}[h]
\setlength{\abovecaptionskip}{0.cm}
\centering
\includegraphics[width = \linewidth]{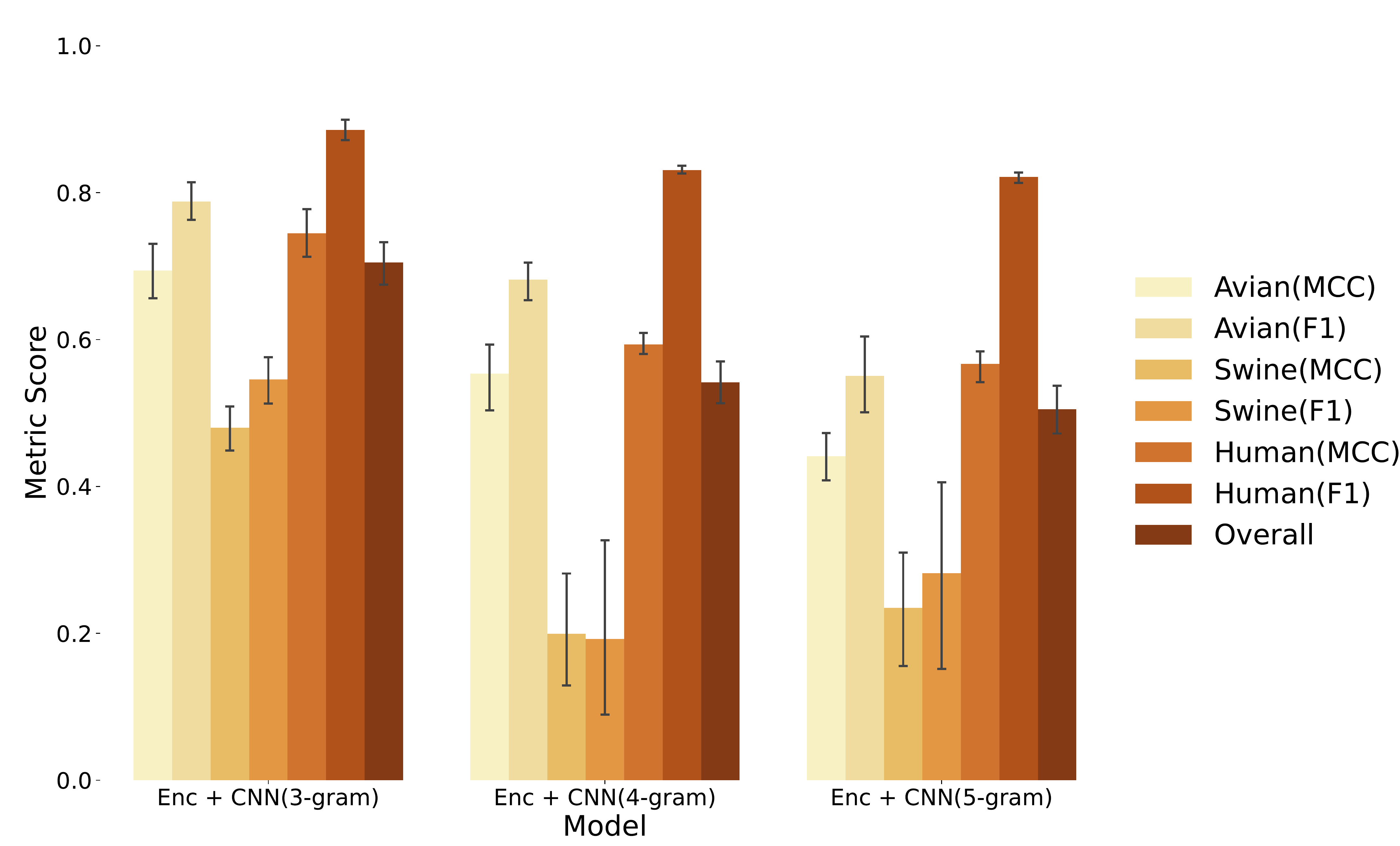}
\caption{\small Performance of Encoding-CNN in Individual Classes}
\label{fig_results_cnn_classes}
\end{figure}

Regardless of the length of the n-gram, the overall performance of word embedding is better than word encoding in CNN (Fig.~\ref{fig_results_cnn_overall}). Human is the easiest class among all CNN-based models to classify, and swine is the most difficult one (Fig.~\ref{fig_results_cnn_classes}).

\begin{figure}[h]
\setlength{\abovecaptionskip}{0.cm}
\centering
\subfloat[\small Encoding-CNN]{%
  \includegraphics[width=0.16\textwidth]{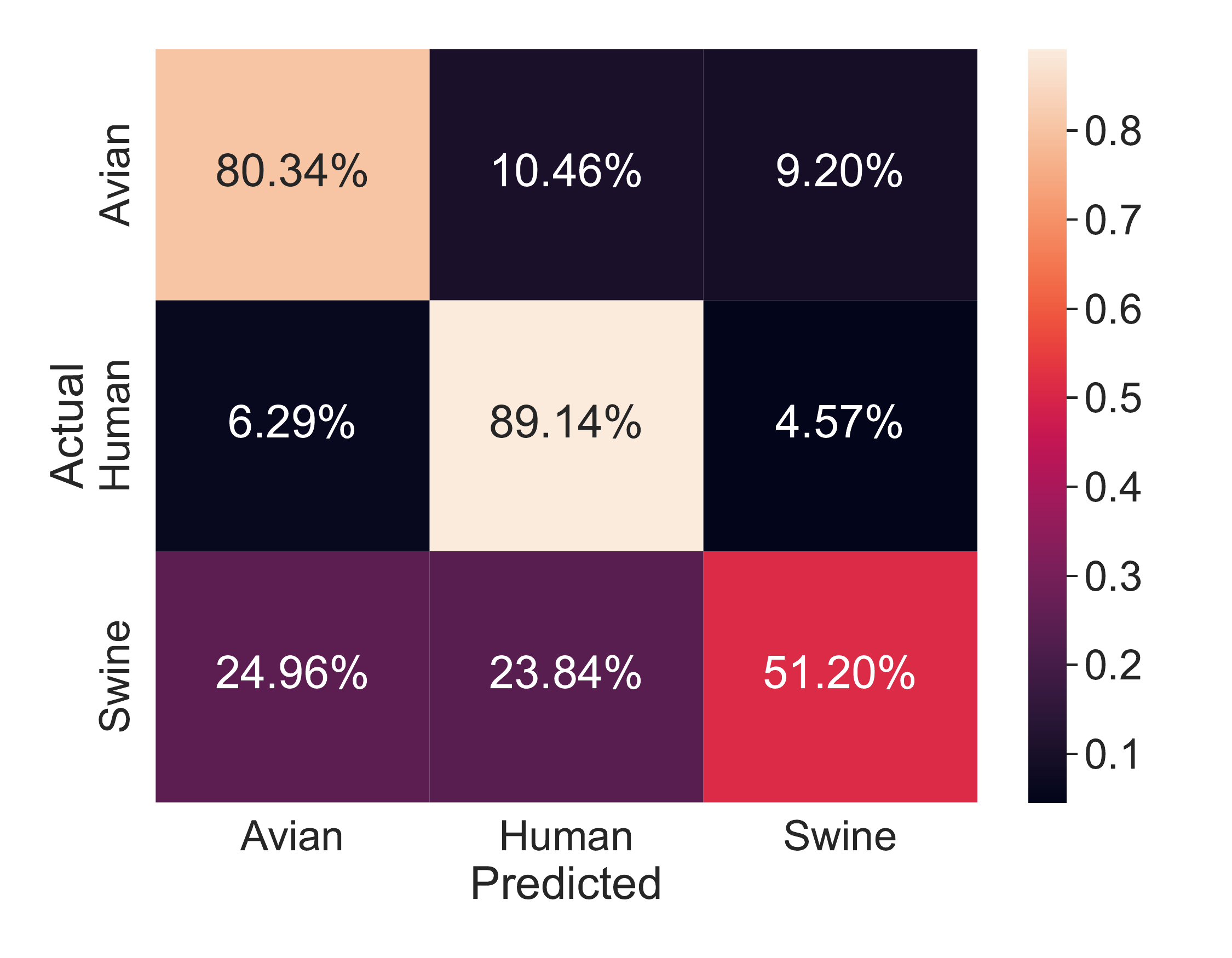}%
  \label{fig_cm_enc}
}
\subfloat[\small Embedding-CNN]{%
  \includegraphics[width=0.16\textwidth]{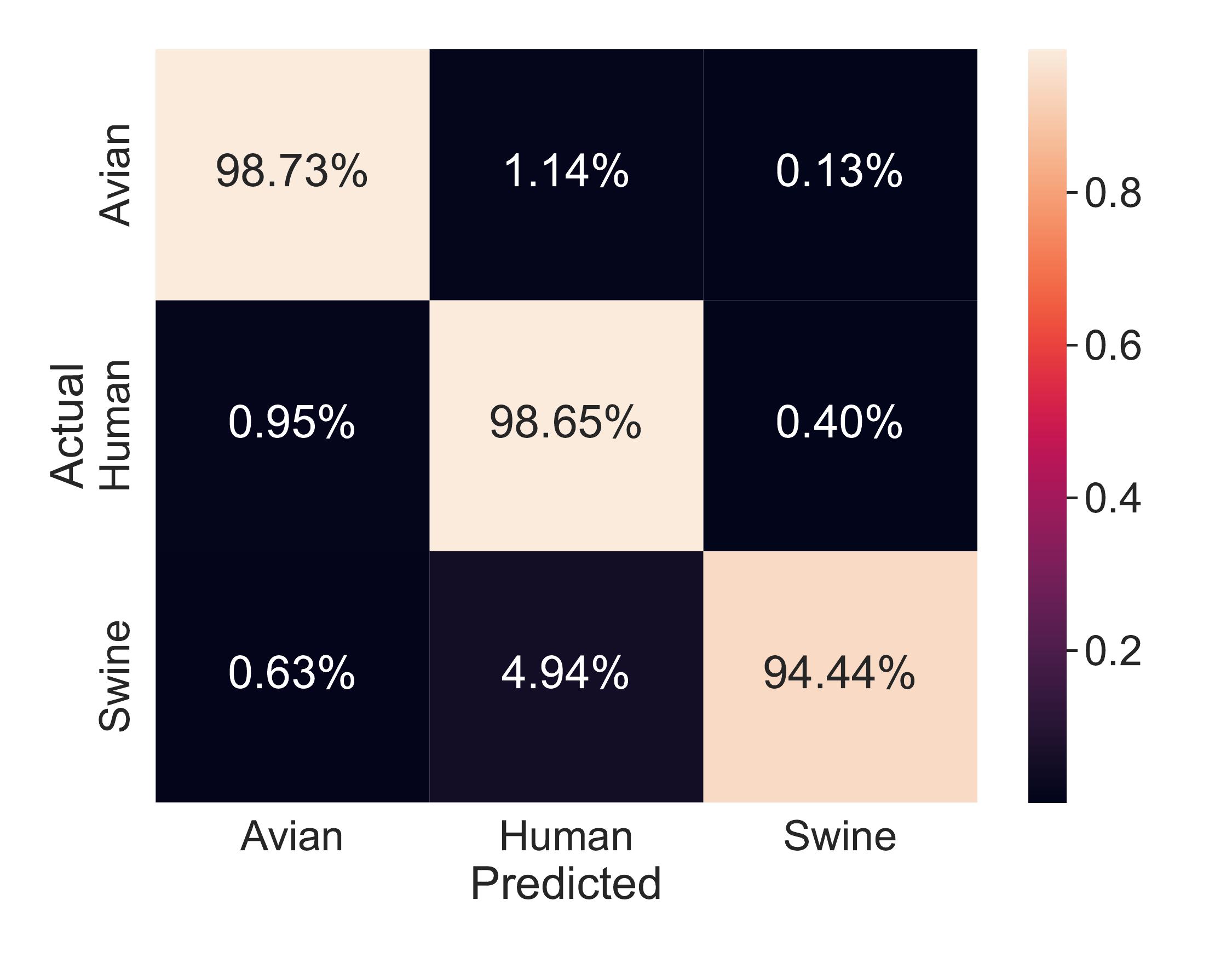}%
  \label{fig_cm_emb}
}
\subfloat[\small PSSM-ML]{%
  \includegraphics[width=0.16\textwidth]{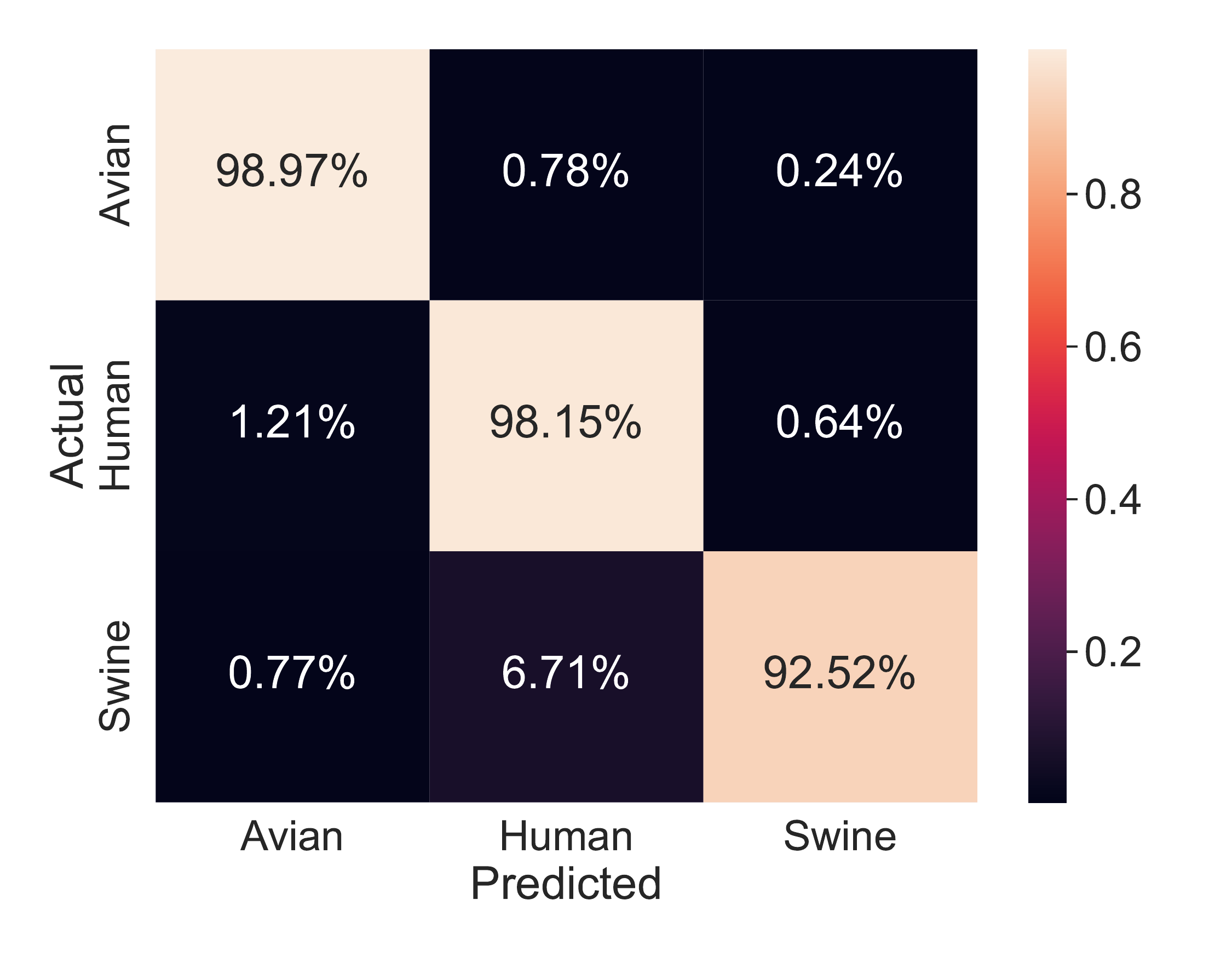}%
  \label{fig_cm_pssm}
}
\caption{\small Confusion Matrix of Different Models: the value on $i$th row and $j$th column indicates the probability of predicting the class $i$ as the class $j$.}
\label{fig_cm}
\end{figure}

Fig.~\ref{fig_cm} depicts the average performance of all PSSM-based optimized machine learning models (Fig.~\ref{fig_cm_pssm}) and CNN-based models (Fig.~\ref{fig_cm_enc}, Fig.~\ref{fig_cm_emb}) in each class based on the confusion matrix on the test set. The human class has above 92\% probability to be correctly classified among all models, as opposed to the swine. The performance between PSSM-ML and Embedding-CNN in human and swine classes is not much different, and Embedding-CNN is better than PSSM-ML in distinguishing swine viruses.

\subsection{Integrated Results from Various models}

We looked at the performance of the models and integrated the predicting results from Embedding-CNN and PSSM-ML. For each viral sequence, we only know the infected host of the sequence instead of the original host. Some sequences were collected during the outbreak, which means that the host used to isolate the virus may not be the original host. Previous research~\cite{add_5} only used one model to predict the viral host, which raises a concern about the accuracy of results. Different models may yield different predictions. Therefore, we provided integrated results and investigated the agreement between the model results, see details in Table~\ref{tab_mis_all_models}.

\begin{table}[h]
\setlength{\abovecaptionskip}{0.cm}
\centering
\caption{\small Integrated Results}
\label{tab_mis_all_models}
\begin{threeparttable}
\includegraphics[width = 3 in]{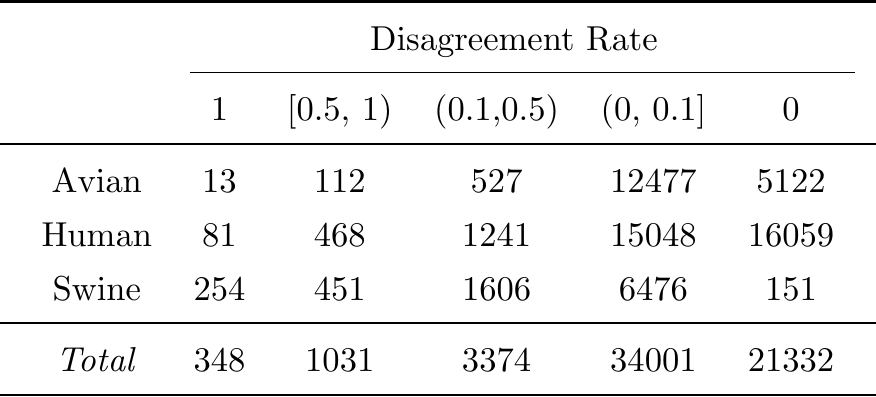}
\begin{tablenotes}
  \item[*] \emph{disagreement rate of sequence $i$ equals to $\nicefrac{E_i}{N}$, where $E_i$ is the number of models that misclassify sequence $i$, and $N$ is the total number of models. If the disagreement rate of sequence $i$ is in $[0.5,1)$, it means that sequence $i$ cannot be correctly classified by more than half of the models.}
\end{tablenotes}
\end{threeparttable}
\end{table}

The models can reach a 100\% agreement of approximate 35.50\% of the sequences, while 100\% disagreement between models occurs in 348 sequences. Among these 348 sequences, 217 belonged to the H1N1 influenza virus, and 196 sequences were similar to or related to the 2009 H1N1 influenza pandemic (pH1N1), including 55 sequences with mixed positive and negative segments. More specifically, 6 out of 13 avian sequences and 190 out of 254 swine sequences, all probably related to the pH1N1. These sequences show the potential ability to break the barrier of species and could be isolated from different species.

From the machine learning perspective, the predicted label mismatch the true label means the model misclassify the data. But this view is open to discussion for Influenza viral host prediction. We listed six sequences that can be traced in literature, as shown in Table~\ref{tab_example}. These sequences seem to be misclassified by all models, but the literature findings gave us the opposite result.

\begin{table}[ht]
\setlength{\abovecaptionskip}{0.cm}
\centering
\caption{\small Example of Strains Reach Model Disagreement}
\includegraphics[width = \linewidth]{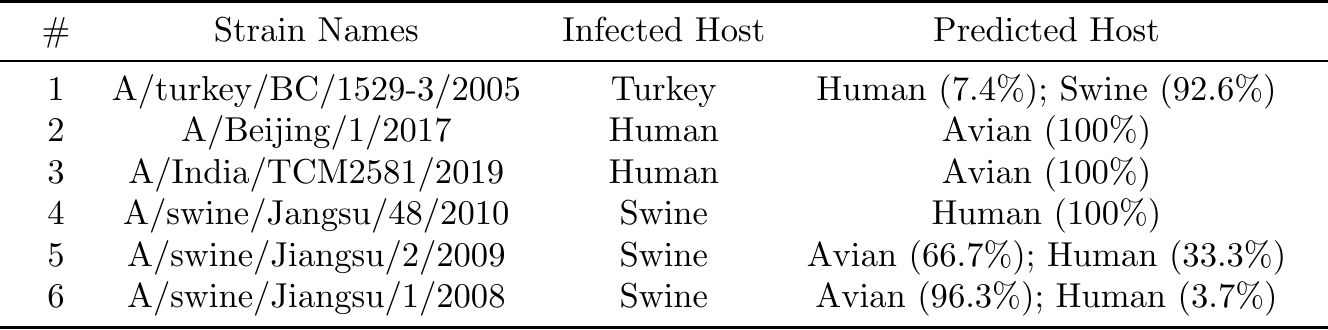}
\label{tab_example}
\end{table}

Previous research speculate that \emph{A/turkey/BC/1529-3/2005} may be a swine-origin virus~\cite{add_7}, \cite{add_12}. \emph{A/Beijing/1/2017} and \emph{A/India/TCM2581/2019} are avian viruses isolated from human~\cite{add_8},~\cite{add_9}. \emph{A/swine/Jangsu/48/2010} is a pH1N1-like swine virus which used for proving the retro-infection from swine to human in China~\cite{add_10}. \emph{A/swine/Jiangsu/1/2008} was isolated from swine but prove to be mostly close relationship to avian~\cite{add_11}. \emph{A/swine/Jiangsu/2/2009} was also isolated from infected swine but also mostly closely related to avian and human~\cite{add_11}.

\subsection{Performance at a Lower Taxonomic Level}
\begin{figure}[h]
\setlength{\abovecaptionskip}{0.cm}
\centering
\subfloat[\small Embedding-CNN (5-gram)]{%
  \includegraphics[width=\linewidth]{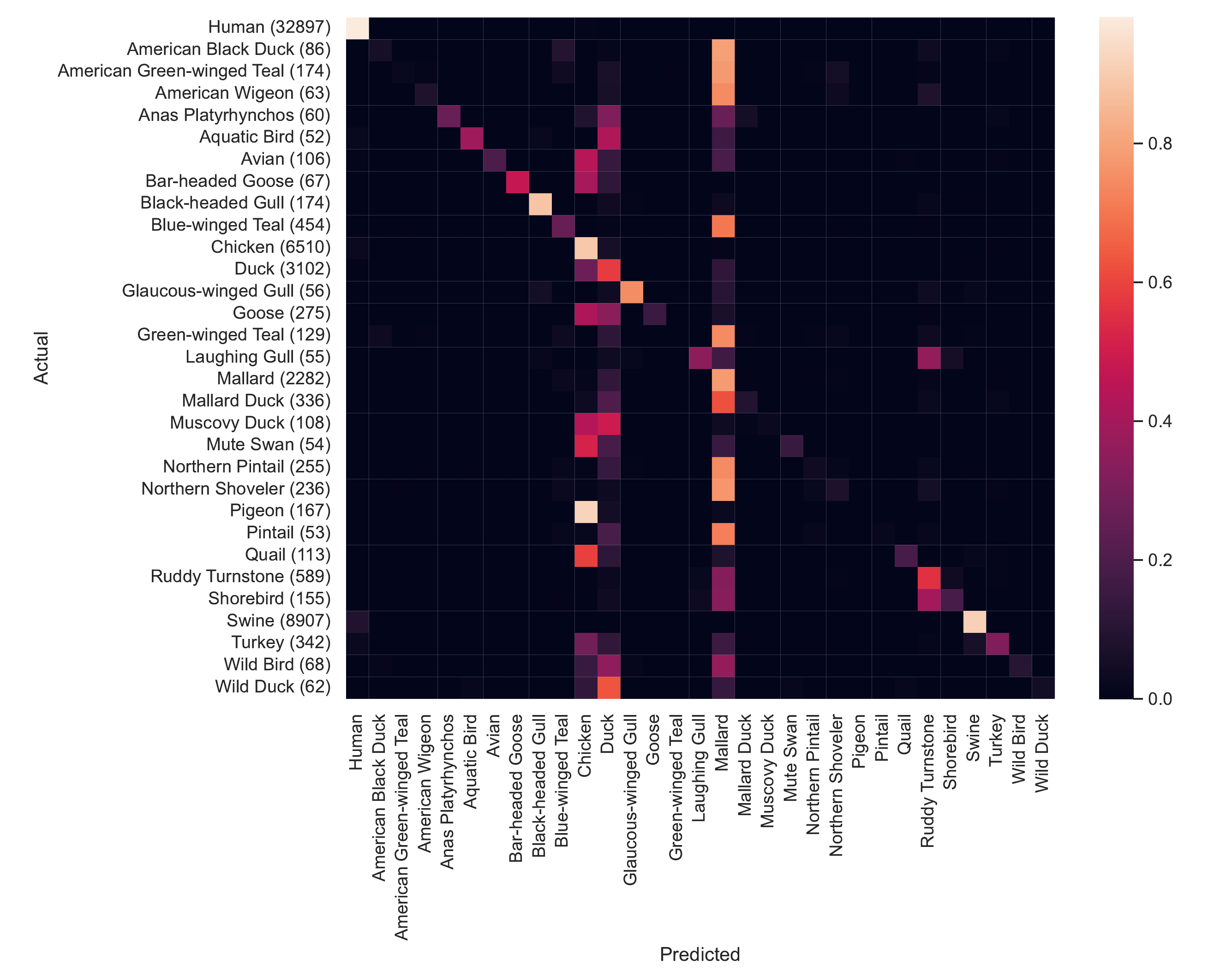}%
}
\hfill
\subfloat[\small ER-PSSM-XGB]{%
  \includegraphics[width=\linewidth]{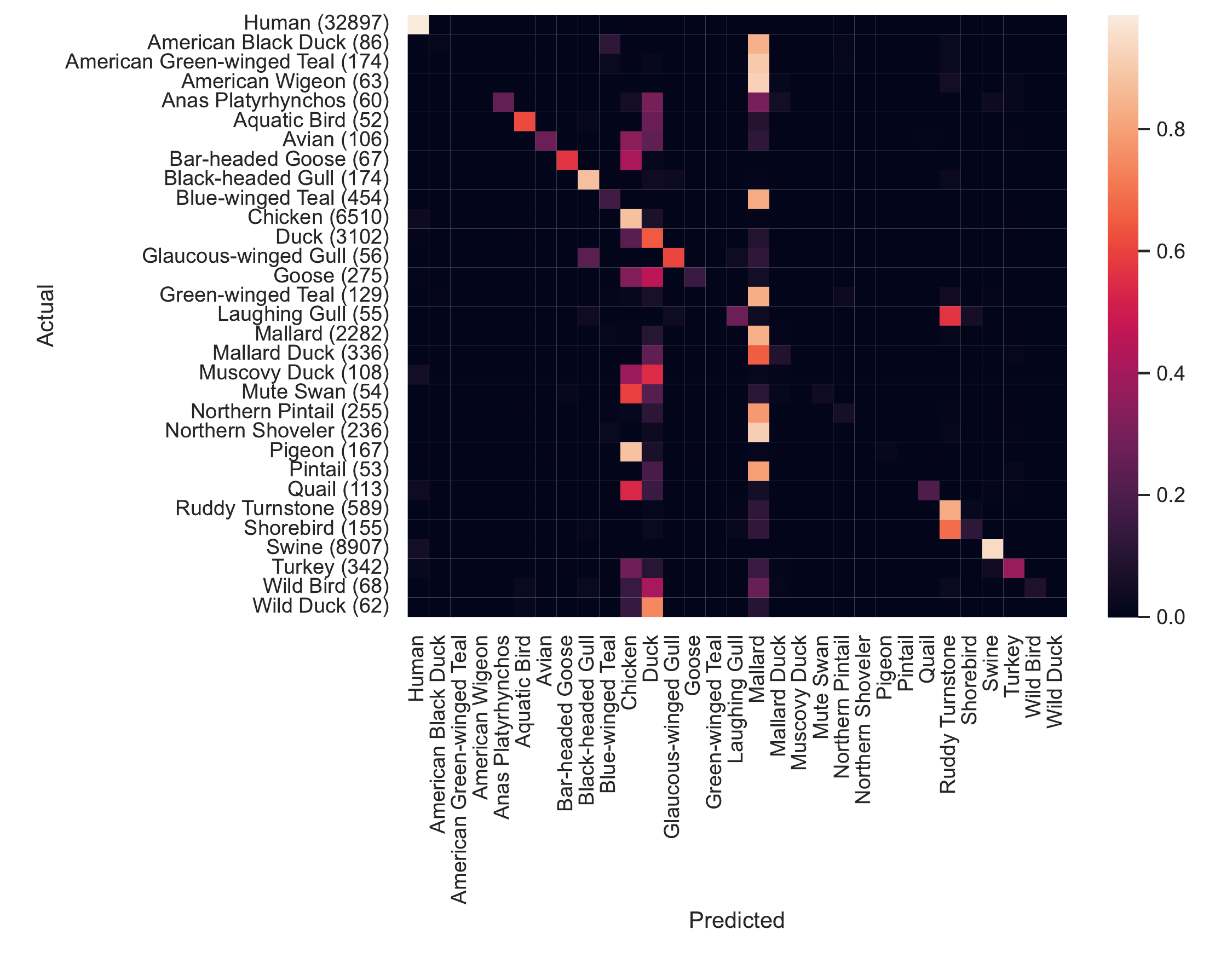}%
}
\caption{\small Confusion Matrix of Embedding-CNN (5gram) and ER-PSSM-XGB at a Lower Taxonomic Level: the number of sequences for each class is indicated in the brackets of actual label.}
\label{fig_lower}
\end{figure}

Above results only represented the performance of proposed models at a higher taxonomic level. This subsection illustrates the performance of models at a lower taxonomic level. Among all PSSM-based models, the performance of ER-PSSM-XGB  is outstanding and stable. As for CNN-based models, Embedding-CNN with 5-gram yields the best results. Hence, we also evaluated the performance of ER-PSSM-XGB and Embedding-CNN (5-gram) when it comes to a lower taxonomic level. Fig.~\ref{fig_lower} shows the confusion matrix of Embedding-CNN (5-gram) and ER-PSSM-XGB at a lower taxonomic level. 

There is not much difference in performance between Embedding-CNN (5-gram) and ER-PSSM-XGB. However, some host can be detected by ER-PSSM-XGB but not Embedding-CNN (5-gram), such as mute swan. The overall performance of Embedding-CNN (5-gram) and ER-PSSM-XGB are both degraded compared with their performance at a higher taxonomic level. Embedding-CNN (5-gram) produces a 89.03\% $F_1$ and a 82.75\% MCC, whilst ER-PSSM-XGB has a 87.63\% $F_1$ and a 80.52\% MCC. In addition to this, the number of sequences correctly classified by ER-PSSM-XGB (87.63\%) is higher than Embedding-CNN (5-gram) (62.18\%).

\section{Discussion and Conclusion}
We used both sequences alignment-based and sequence alignment-free models to predict viral hosts of influenza A viruses. Sequence alignment-based models use the features extracted by PSSM before feeding into classic machine learning models. PSSM-based features cannot only maintain the evolutionary information in the virus sequence but also capture the local information in the sequence. Sequence alignment-free models transfer the virus sequences into numerical vectors through word encoding or word embedding before feeding into CNN and is incapable of capturing the local information in sequences. They also need padding to make the dimension of the vectors uniform. But even so, the overall performance of Embedding-CNN is  still slightly better than PSSM-based models at a higher taxonomic level. When it comes to the performance at a lower taxonomic level, the PSSM-based model can classify some viruses that Embedding-CNN cannot do.

The data set used in the paper is non-redundant and all the multi-label sequences were removed, which means each sequence only belongs to a single host. Multi-label prediction is beyond the scope of this article (only about 3\% of the data is multi-label) and is left for future work. 

\section*{Acknowledgment}
This work is supported by the University of Liverpool.

\bibliographystyle{unsrt}

\end{document}